\pdfoutput=1
\documentclass[11pt]{article}

\newif\ifarxiv
\arxivtrue % make it 'true' or 'false' if you want to keep the anonymized or public version

\ifarxiv
    \usepackage{acl2023}
\else
    \usepackage[review]{acl2023}
\fi
\usepackage{times}
\usepackage{latexsym}

\usepackage[T1]{fontenc}
\usepackage[utf8]{inputenc}

\usepackage{microtype}

\usepackage{inconsolata}

\usepackage{graphicx}

\usepackage[capitalise]{cleveref}
\usepackage{caption}
\usepackage{subcaption}
\usepackage{multirow}
\usepackage{enumitem}
\usepackage{todonotes}
\usepackage{adjustbox}
\usepackage[normalem]{ulem} % [normalem] prevents the package from changing the default behavior of `\emph` to underline.

\usepackage{colortbl}

\crefname{figure}{Figure}{Figures}
\crefname{table}{Table}{Tables}
\crefname{appendix}{Appendix}{Appendices}
\crefname{section}{Section}{Sections}
\crefname{equation}{Eq.}{Eqs.}

\definecolor{pink}{RGB}{219, 48, 122}
\definecolor{forestgreen}{RGB}{34,139,34}
\definecolor{goldenrod}{RGB}{218,165,32}
\definecolor{sepia}{RGB}{112,66,20}
\newcommand\myparagraph[1]{
\vskip 0.05in 
\noindent{\bf {#1}}}
\title{
When Does Aggregating Multiple Skills with Multi-Task Learning Work?\\A Case Study in Financial NLP
}

\author{Jingwei Ni \\
  ETH Zürich \\
  \texttt{njingwei@ethz.ch} \\\And
  Zhijing Jin \\
  MPI \& ETH Zürich \\
  \texttt{jinzhi@ethz.ch} \\\AND
  Qian Wang \\
  University of Zürich \\
  \texttt{qian.wang@uzh.ch} \\\And
  Mrinmaya Sachan \\
  ETH Zürich \\
  \texttt{msachan@ethz.ch} \\\And
  Markus Leippold \\
  University of Zürich \& SFI\\
  \texttt{markus.leippold@bf.uzh.ch} \\
  }

\begin{document}
\maketitle
\begin{abstract}
Multi-task learning (MTL) aims at achieving a better model by leveraging data and knowledge from multiple tasks. %MTL can help in domains where we have multiple small datasets catered to different skills but few large datasets. 
However, MTL does not always work -- sometimes negative transfer occurs between tasks, especially when aggregating loosely related skills, leaving it an open question when MTL works. %making it unclear when MTL is effective. % This question has been partially answered by previous work from an algorithmic perspective. However, practical reasons are less well understood.
Previous studies show that MTL performance can be improved by algorithmic tricks. However, what tasks and skills should be included is less well explored.
%However, previous work has not explored this question from the perspective of task aggregation, especially the scenario where multiple skills are aggregated.
In this work, we conduct a case study in Financial NLP where multiple datasets exist for skills relevant to the domain, such as numeric reasoning and sentiment analysis. Due to the task difficulty and data scarcity in the Financial NLP domain, we explore when aggregating such diverse skills from multiple datasets with MTL can work.
Our findings suggest that the key to MTL success lies in skill diversity, relatedness between tasks, and choice of aggregation size and shared capacity.
% are all important factors in achieving successful MTL performance. 
Specifically, MTL works well when tasks are diverse but related, and when the size of the task aggregation and the shared capacity of the model are balanced to avoid overwhelming certain tasks.\footnote{
\ifarxiv
\url{https://github.com/EdisonNi-hku/MTL4Finance}.
\else
We will open-source our codes upon acceptance.
\fi
}
\end{abstract}

\section{Introduction}
%\mrinmaya{I have a suggest for a small re-framing of our contribution. Maybe we want to claim that this paper is contributing an interesting analysis on when/how aggregation of multiple skills can happen in a single model with MTL (with Fin NLP as a case study). And if there can be challenges with this and this may not work. I am wondering if readers will find the aggregation for finance NLP a big contribution as there are probably better FinNLP models than us e.g. FinBERT like work.}

Multi-task learning (MTL) is a machine learning paradigm where multiple learning tasks are optimized simultaneously, exploiting commonalities and differences across them \cite{Caruana1997MultitaskL}. MTL is expected to outperform single-task learning (STL) as it utilizes more training data and enables inter-task knowledge sharing \citep{ruder2017mtl}. However, MTL may also bring about multi-task conflict and negative transfer. Empirically, in many MTL systems, only a small portion of tasks benefit from MT joint training while others suffer from negative transfer \citep{stickland2019pal,raffel2020t5,peng-etal-2020-empirical}. Therefore, it is still an open question \textit{when MTL will work}.

\begin{table}[t]
\centering
\begin{tabular}{lcc}
\hline
\textbf{Method}             & \textbf{TSA$\downarrow$}   & \textbf{SC$\uparrow$}                 \\ \hline
GPT-3 Zero-Shot  & $0.3700$          & $77.69\%$          \\
GPT-3 Few-Shot  & $0.3128$          & $80.37\%$          \\
FinBERT Fine-Tune & $\mathbf{0.2054}$ & $\mathbf{86.61\%}$ \\ \hline
\end{tabular}
\caption{
Performance comparison of our method (FinBERT Fine-Tune) and GPT-3 (text-davinci-003) baselines. We report the rooted mean square error ($\downarrow$) on the task of target-based sentiment analysis (TSA) \citep{cortis-etal-2017-semeval} and accuracy ($\uparrow$) on sentiment classification (SC) \citep{financialphrasebank2013malo}.
% TSA denotes target-based sentiment analysis \citep{cortis-etal-2017-semeval}, for which we report rooted mean square error; SC denotes sentiment classification \citep{financialphrasebank2013malo}, for which we report accuracy. 
See GPT-3 prompts and settings in \cref{appendix:gpt3}.
}\label{tab:gpt-3}

\end{table}
%\mrinmaya{Is this table cited in the intro? It should be (); otherwise, its relevance is not clear.}

MTL systems have two components: MTL algorithms and the tasks included for aggregation. Recent progress in MTL has shown that appropriate MTL algorithms (e.g., architecture and optimization) can mitigate negative transfers (\citealp{gradient2020yu,wang2021gradient,pfeiffer-etal-2021-adapterfusion,karimi-mahabadi-etal-2021-parameter,mao-etal-2022-metaweighting,ponti2022combining}, \textit{inter alia}). %\citep{khashabi-etal-2020-unifiedqa} shows that concentrating on a single skill (question answering) benefits MTL, but the scenario where multiple skills are aggregated is under-explored.
However, it is still unclear when MTL works from the perspective of the relations between tasks and skills to be aggregated for better performance in a practical setting.

To understand this, we 
%To analyze these hypotheses, we 
conduct a practical case study on Financial NLP. We choose Financial NLP mainly because (1) Financial NLP tasks are hard: GPT-3 \citep{gpt-32020} does not perform well on financial tasks (see \cref{tab:gpt-3}), though it is a good zero/few-shot learner in general domains; and (2) Financial NLP datasets typically address different skills (e.g.,  quantitative reasoning, and sentiment analysis), and have a limited data size (\citealp{financialphrasebank2013malo,cortis-etal-2017-semeval,lamm-etal-2018-textual,mariko-etal-2020-financial,chen2020overview,chen2022overview}, \textit{inter alia}). Therefore, it is promising to aggregate Financial NLP tasks using MTL, which not only compiles and augments the small datasets, but also benefits the difficult tasks through relevant information transfer and comprehensive reasoning. However, no previous work explores the benefits of aggregating Financial NLP resources using MTL. Particularly, we explore the following \textit{\textbf{hypotheses}} about when MTL works: 
\begin{enumerate}[itemsep=0pt,topsep=1pt,label=\textit{H\arabic*.}]
    \item \textit{\textbf{When various skills are included}}: Intuitively, positive transfers are likely to happen among tasks regarding the same skill. %\citet{khashabi-etal-2020-unifiedqa} also shows that MTL often beats STL when concentrating on a single skill (question answering).
    %However, tasks corresponding to different NLP skills might also have good transferability to each other. Furthermore, 
    However, diversified skills might benefit the MTL system through implicit data augmentation, attention focusing, and feature eavesdropping \cite{ruder2017mtl}. Our empirical results also show that skill diversity benefits MTL. % but the degree of skill diversity should be appropriate to reduce negative transfer.
    \item \textit{\textbf{When the aggregated tasks are well related}}: % how important is task-relatedness to the success of MTL? %How to measure task-relatedness?
    %While diverse skills may benefit MTL, tasks that are too distantly related can still cause negative transfers.
    We find that the close relation (measured qualitatively and quantitatively) among Financial NLP tasks explains why diversified skills help each other, and contributes to the 
    % explains why MTL works.
    % Therefore, task-relatedness is also important to the 
    success of MTL.
    \item \textit{\textbf{When the aggregation size matches shared capacity}}: Too many objectives may exhaust the MTL shared capacity and cause interference among tasks \citep{stickland2019pal}. 
    We find that excessive aggregation size in a limited capacity model restricts the performance of some tasks. Thus aggregation size should be appropriate for the shared capacity.
\end{enumerate}

To facilitate exploration of \textit{H1} and \textit{H2}, we survey existing Financial NLP resources and propose \textbf{FinDATA} (\underline{\textbf{Fin}}ancial \underline{\textbf{D}}ata \underline{\textbf{A}}nd \underline{\textbf{T}}asks \underline{\textbf{A}}ggregation), a collection of Financial NLP tasks covering various financial text understanding skills. To check \textit{H3}, we propose \textbf{SPAL-FinBERT} (\underline{\textbf{S}}hared \underline{\textbf{P}}arallel \underline{\textbf{A}}ttention \underline{\textbf{L}}ayer with \underline{\textbf{FinBERT}}), an MTL architecture based on pre-trained FinBERT \cite{araci2019finbert}, but is highly parameter-efficient -- with $99.8\%$ fewer trainable parameters but outperforming the vanilla FinBERT MTL on several tasks.
% with tunable capacity. SPAL-FinBERT is also parameter-efficient: it outperforms vanilla FinBERT MTL on several tasks, with $99.8\%$ fewer trainable parameters. 
Our contributions include
%\mrinmaya{Can this be merge into the hypotheses above to save space?}
%Through our case study we find that: (1) tasks with different NLP skills can positively transfer to each other, and some degree of skill diversity contributes to the success of MTL. (2) The relatedness of Financial NLP tasks explains why MTL works for financial domain, and the relatedness can be measured qualitatively and quantitatively. (3) Aggregating too many skills in limited capacity restricts the performance of some tasks, but does not significantly affect others. Thus aggregation size should be appropriate for both shared capacity and the target task. 
\begin{enumerate}[itemsep=0pt,topsep=1pt
% ,label=C\arabic*.
]
    \item We conduct a case study on Financial NLP to explore what properties of task aggregation lead to the success of MTL.
    \item We survey and aggregate several existing Financial NLP tasks and datasets, illustrating that MTL can be a cheap and efficient improvement for Financial NLP performance.
    \item We propose SPAL-FinBERT, a parameter-efficient MTL architecture with good performance. This model may also have broader use cases in other settings.
\end{enumerate}

\section{Background \& Related Work}
%\subsection{Multi-task Learning}
Previous work mainly focuses on two categories of MTL practice: MTL as pre-training and MTL as auxiliary training. 

\myparagraph{MTL as pre-training:} Besides unsupervised pre-training, supervised data can also be utilized for pre-training in an MTL manner (i.e., an \textit{intermediate} training stage) to improve the model's multi-aspect intelligence and generalizability to unseen tasks. Such an approach has been shown beneficial for various pre-trained models, including encoder-only models \citep{liu-etal-2019-multi,aghajanyan-etal-2021-muppet}, encoder-decoder models \citep{aribandi2022ext,chung2022scaling}, and large language models \citep{wei2022finetuned,Sanh2021t0,min-etal-2022-metaicl,chung2022scaling}. \citet{aghajanyan-etal-2021-muppet} show that MTL pre-training does not work with small-scale task aggregation. More recent analysis shows that aggregating related tasks transfers better to a known target task \citep{padmakumar-etal-2022-exploring}. 

\myparagraph{MTL as auxiliary training:} Instead of training a target task alone, we can jointly train it with other auxiliary tasks to improve its performance in an MTL manner (i.e., the \textit{final} training stage). However, this approach does not work in most cases, especially when multiple skills are aggregated (e.g., GLUE) \citep{stickland2019pal,peng-etal-2020-empirical,raffel2020t5,mueller2022t2t}. %When MTL auxiliary training focus on one skill (e.g. question answering), larger portion of included tasks outperforms STL \citep{khashabi-etal-2020-unifiedqa}. 
Previous work shows that appropriate algorithmic tricks lead to more success in MTL: (1) MTL architecture: \citet{pfeiffer-etal-2021-adapterfusion,karimi-mahabadi-etal-2021-parameter} and \citet{ponti2022combining} propose MTL architectures that encourage high-level knowledge sharing instead of direct parameter-sharing; and (2) MTL optimization: \citet{gradient2020yu} and \citet{wang2021gradient} geometrically manipulate the gradients to reduce the conflicts, and \citet{mao-etal-2022-metaweighting} learn to weight losses of including tasks automatically. In computer vision (CV), \citet{fifty2021efficiently} address that task aggregation is also crucial for MTL besides algorithms and proposes an algorithm to select the best task grouping from a task collection. However, rare previous work in NLP analyzes what task aggregation leads to the success of MTL and what qualities of a task aggregation are important. 

\section{FinDATA Compilation}
We compile FinDATA, a task aggregation on Financial NLP, to facilitate the case study. We first set the desiderata, and then survey existing Financial NLP tasks to select those that meet these criteria.

\subsection{Desiderata}
\myparagraph{Diversified skills:} We are interested in the importance of skill diversity and task-relatedness in MTL. Therefore, included tasks should cover as many Financial NLP skills as possible. If multiple tasks correspond to the same skill (e.g., sentiment analysis), we prefer smaller ones that are more worth aggregating and less likely to dominate. Some tasks can have closer relation than others (e.g., corresponding to similar skills).

\myparagraph{Aligned form of input:} To enable joint training, we prefer tasks with sentences or paragraphs as inputs, instead of phrases, tables, or full reports.

\subsection{Financial NLP} \label{sec:fin_nlp}
The most prevalent Financial NLP task is sentiment analysis on financial tweets or news, as it directly contributes to automatic decision-making tools in the financial market. There are two types of financial sentiment analysis, the first of which defines sentiment analysis as a coarse-grained classification problem. Given a piece of financial news, the system only needs to classify its sentiment into positive, negative, or neutral. Most of the financial sentiment analysis are in this form, for example, Financial Phrase Bank \citep{financialphrasebank2013malo}, and StockSen \citep{xing-etal-2020-financial}. The other instantiation of financial sentiment analyses has more fine-grained labels: \citet{cortis-etal-2017-semeval} assigns different sentiment scores from $-1$ to $1$ to different targets in financial news. %For example, in "Clinigen chosen by AstraZeneca to manage access programme for next...", the sentiment score corresponding to "Clinigen" is $0.328$ (positive) while the sentiment score of "AstraZeneca" is $0.024$ (neutral). FiQA\footnote{https://sites.google.com/view/fiqa/home} further adds financial aspect labels to a part of this target-based sentiment analysis dataset.

%Besides, numbers are ubiquitous in all forms of financial text (e.g., news, tweets, and reports). Hence, many tasks and datasets are proposed for number semantics and numeracy. For example, number understanding \citep{chen2019overview,chen2022overview}, number attachment prediction \cite{chen2020overview}, magnitude prediction \cite{chen-etal-2019-numeracy}, and Q/A regarding financial tables and numeracy \cite{zhu-etal-2021-tat,chen-etal-2021-finqa}.
Numbers are ubiquitous in all forms of financial text (e.g. news, tweets, and reports). Hence, many tasks and datasets are proposed for number semantics and numeracy. For example, FinNum shared task of recent years proposed several datasets focusing on financial number type understanding and number attachment \citep{chen2019overview,chen2020overview,chen2022overview}. \citet{chen-etal-2019-numeracy} further proposed Numeracy-600K for number magnitude understanding. \citet{zhu-etal-2021-tat} proposed TAT-QA, a Question Answering(QA) benchmark financial hybrid (tabular and text) data. Similarly, \citet{chen-etal-2021-finqa} proposed FinQA, another QA benchmark on financial hybrid data emphasizing numeracy skills.

%Some datasets provide financial natural language understanding (NLU) skills other than sentiment and numbers: for instance, \citet{lamm-etal-2018-textual} with financial semantic role labels and \citet{mariko-etal-2020-financial} for causality detection in financial news. Furthermore, there are tasks that take entire documents as inputs, for example, narrative summarization  \citep{el-haj-etal-2020-financial} and table of content prediction \citep{maarouf-etal-2021-financial}. Some other tasks focus on financial concepts \citep{maarouf-etal-2020-finsim,kang-etal-2021-finsim,finsim4}. \cref{sec:all_datasets} covers more details regarding mentioned datasets.
Some datasets provide financial natural language understanding (NLU) skills other than sentiment and numbers. For instance, \citet{lamm-etal-2018-textual} proposed a dataset for analogy parsing originally, which contains financial semantic role annotations and thus can be used for semantic role labeling (SRL). \citep{mariko-etal-2020-financial} detects causal effect in financial news. 

Not all financial NLP tasks are sentence-level. Many tasks take entire documents as inputs, for example, narrative summarization  \citep{el-haj-etal-2020-financial} and  table of content prediction \citep{maarouf-etal-2021-financial} on financial reports. Some other tasks focus on financial concepts (phrases) \citep{maarouf-etal-2020-finsim,kang-etal-2021-finsim,finsim4} instead of complete sentences. \cref{sec:all_datasets} covers more details regarding mentioned datasets.

\subsection{FinDATA}
\begin{table*}[t]
\centering
%\small
\centering
\resizebox{\textwidth}{!}{
\begin{tabular}{lcccllll}
\hline
 & \textbf{|Train|} & \textbf{|Dev|} & \textbf{|Test|} & \textbf{Task Type} & \textbf{Metrics} & \textbf{Text Source} & \textbf{Financial NLU Skill}\\
\hline
TSA & 913 & 229 & 561 & Seq Regression & RMSE & News texts  & Sentiment w.r.t. target\\
SC & 3866 & 484 & 484 & Seq Classification & Accuracy & News texts & Sentiment type\\
NC & 6669 & 1668 & 1191 & Seq Classification & Accuracy & Analyst report & Number type \\
NAD & 7187 & 1044 & 2109 & Seq Classification & Accuracy & Financial tweets & Number attachment \\
FSRL & 900 & 100 & 100 & Token Classification & Macro-F1 & News texts & Semantic roles \\ 
CD & 674 & 226 & 226 & Span Prediction & Accuracy & News texts & Causal effect\\\hline
\end{tabular}
}
\caption{
\label{tab:findata} Statistics of all FinDATA tasks and datasets. We report the sizes of train, development, and test splits. If there is no official test or development set, we split the training set by ourselves (more details in \cref{sec:data_split}).}
%\vspace{-0.5em}
\end{table*}
%Based on our survey and desiderata, the following tasks (statistics in \cref{tab:findata}) are selected: 
%There are two predominant skills in Financial NLP, namely sentiment analysis and financial number understanding since a large portion of existing datasets is focusing on these skills. %(7 out of 17 datasets are number related; 4 out of 17 datasets are sentiment related, see \cref{tab:all_stat}). 
%We also include two other skills: semantic understanding and causality awareness. 
Based on our survey and desiderata, the following 4 Financial NLP skills are selected:

%\mrinmaya{I'd suggest spending a bit more space on this section and writing each skill, describing it and its datasets in one paragraph. I would shorten the related work (at least in the submission version) if pressed for space.}
%\mrinmaya{I liked the earlier version of this with the desiderata, etc.}

%As a result, for sentiment, we select 
\myparagraph{Financial sentiment analysis} is a prevalent skill in the Financial NLP domain, analyzing financial news' and investors' sentiment toward particular financial objects. We select two tasks for this skill: (1) Financial Phrasebank sentiment classification (SC, \citealp{financialphrasebank2013malo}): given a financial news headline, classifying it into positive, negative, or neutral; and (2) SemEval-2017 target-based sentiment analysis (TSA, \citealp{cortis-etal-2017-semeval}): predicting a sentiment score between -1 and 1 w.r.t. a financial news headline and a target company.

%\myparagraph{SemEval-2017 target-based sentiment analysis} (TSA, \citealp{cortis-etal-2017-semeval}): given a financial news headline and a target company, the model should predict a sentiment score between -1 and 1 for the target company.

%For number understanding, we select 
\myparagraph{Financial number understanding} is another important Financial NLP skill, as numbers are ubiquitous in all forms of financial text (e.g., news, tweets, and reports). We select two tasks for this skill: (1) FinNum-3 number classification (NC) \citep{chen2022overview}: given a report paragraph and a target number, classifying it into monetary, percentage, temporal, and so on; and (2) FinNum-2 number attachment detection (NAD) \citep{chen2020overview}: given a financial tweet, a target number, and a cash tag, predicting whether the number is attached (i.e., related) to the cash tag.
%(NC, \citealp{chen2022overview}): given a report paragraph and a target number, the model should understand the number semantic and classify it into monetary (money or change), percentage (relative or absolute), temporal (date or time), quantity (relative or absolute), product number, ranking, or other.

%\myparagraph{FinNum-2 number attachment detection} (NAD, \citealp{chen2020overview}): given a financial tweet, a target number, and a cash tag, the model should predict whether the number is attached (i.e., related) to the cash tag. 

\myparagraph{Financial semantic role labeling} (FSRL) is a skill aiming at understanding the quantitative semantic roles \citep{Lamm2018QSRLA} such as quantity, value, location, date, theme, etc. We include \citeposs{lamm-etal-2018-textual} dataset\footnote{The dataset is for analogy parsing originally. We utilize its quantitative semantic role labels.} for this skill.

\myparagraph{Causality understanding} aims at understanding the causal relationship between financial facts. For this skill, we include FinCausal 2020 Causality Detection (CD, \citealp{mariko-etal-2020-financial}).
%For semantic understanding and causality awareness, we select the financial semantic role labeling (FSRL, \citealp{lamm-etal-2018-textual}) and FinCausal causality detection (CD, \citealp{mariko-etal-2020-financial}) correspondingly. The selected tasks differ in objective even if they are corresponding to a same skill (e.g. TSA is target-based while SC is not). We also exclude datasets with non-sentence input (e.g. documents, phrases, and tables). %Statistics and other details of FinDATA tasks are in \cref{tab:findata}. 

All our datasets are in English. Other details of included tasks can be found in \cref{tab:findata}. We present several examples for each FinDATA dataset in \cref{sec:findata_examples}.

%\mrinmaya{Instead of experiments directly, we can have sections as MTL architectures and Analysis... We can also have more descriptive section names akin to the questions we are answering, e.g., ``Does skill diversity matter?'', etc. which will make things more clear for the reader.}
\section{Multi-Task Learning Systems}
\begin{figure}[t]
    \centering
    \includegraphics[width=\columnwidth]{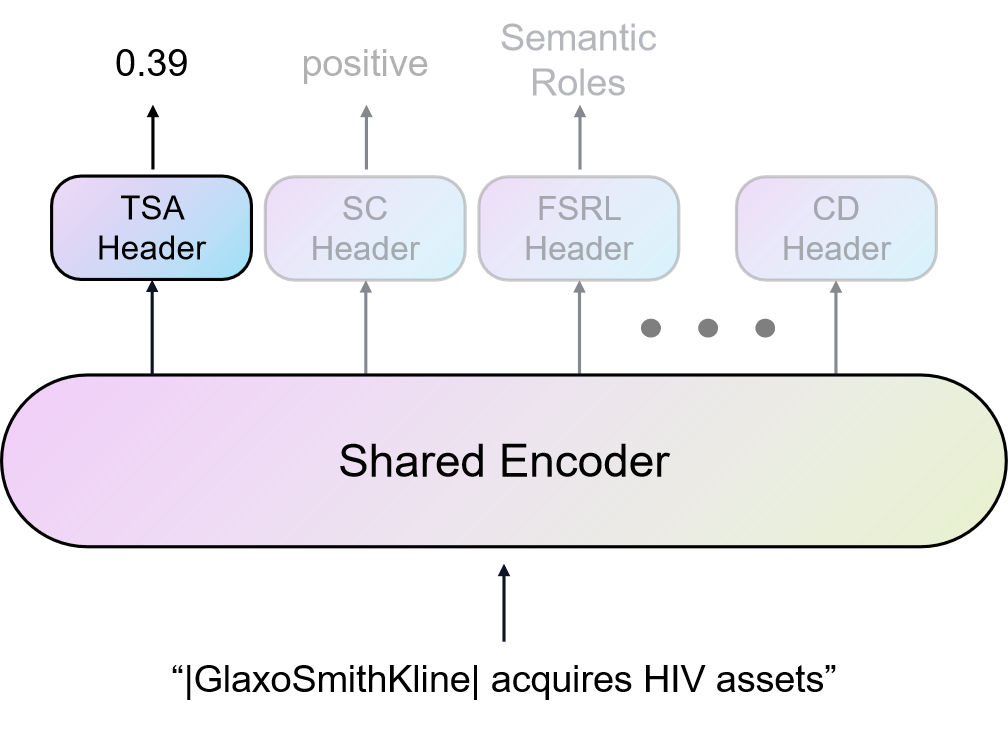}
    \caption{An illustration of MTL system with shared encoder and task-specific prediction headers.}
    %\vspace{-0.7em}
    \label{fig:mtlwithencoder}
\end{figure}

%\myparagraph{MTL Architectures}: 
We consider various MTL systems in the form of shared encoder + one-layer task-specific prediction headers. The MTL problem is formulated as follows:

We are given a joint dataset of multiple tasks $\mathbf{D}=\{(\mathbf{X}_t, Y_t)\}_{t \in \mathbf{T}}$ where $\mathbf{X}_t, Y_t$ denotes the training corpus and labels of task $t$; and $\mathbf{T}$ denotes the task collection. We are also given a pre-trained encoder (e.g., FinBERT) $f_{\theta_E}(\cdot)$ and task-specific prediction headers $h_{\theta_t}(\cdot)$, which are parameterized by $\mathbf{\theta}=(\theta_E, \{\theta_t\}_{t \in \mathbf{T}})$. The training loss for multi-task fine-tuning:
\begin{equation}
\mathcal{L}(\mathbf{\theta}, \mathbf{D})=\sum_{t\in \mathbf{T}} w_{t} \cdot l_{t} (h_{\theta_t} (f_{\theta_E} (\mathbf{X}_t)), Y_t)
%\vspace{-0.7em}
\end{equation}
Where $l_t$ denotes the loss function for task $t$, and $w_{t}$ denotes the sampling weight of task $t$. The generic architecture is illustrated in \cref{fig:mtlwithencoder}.  During training, a task is sampled for each training step, and the corresponding prediction header and the shared encoder are updated (e.g., the TSA example in \cref{fig:mtlwithencoder}).

\section{Experimental Setup}
%In this section, we first introduce our experimental settings, pre-trained models, and MTL architectures. Then we discuss our explorations on qualities of task aggregation that lead to positive transfer.
We fine-tune all MTL and STL models on corresponding data for 40 epochs. (Our MTL batching scheme is described in \cref{sec:experimental_details}.) For STL, we evaluate the model every 50 steps and save the checkpoint with the best validation score. For MTL, we evaluate every 200 steps, saving and reporting the best checkpoint for each task independently following the setting of \citet{raffel2020t5} (i.e., each task can be viewed as the target task with others being auxiliary tasks). % We do not follow \citet{karimi-mahabadi-etal-2021-parameter} to record the checkpoint with the best average validation score because of two reasons: (1) the target score of TSA is Rooted Mean Square Error (RMSE), which can not be averaged with accuracy scores and F1 scores; and (2) our objective is to explore how much each task can benefit from others, instead of obtaining a MTL system balanced at all tasks. 
We follow the evaluation metrics in \cref{tab:findata} to select the best checkpoints and report the test performance.
% by RMSE for TSA, accuracy score for SC, NC, NAD, and CD, and Macro-F1 score for FSRL. The same metrics are computed and reported when testing.
All MTL and STL results are averaged over random seeds from 1 to 5 with standard deviations attached.
\cref{sec:experimental_details} contains more details about data preprocessing, hyperparameters, and GPU usage. 

% \begin{table*}[t]
% \centering
% \begin{tabular}{lcccccc}
% \hline
% STL Models & TSA$\downarrow$ & SC & NC & NAD & FSRL & CD \\
% \hline
% BERT-cased & $0.2320{\scriptstyle \pm0.0082}$ & $86.57{\scriptstyle \pm0.8}$ & $86.19{\scriptstyle \pm1.1}$ & $85.43{\scriptstyle \pm0.8}$ & $71.30{\scriptstyle \pm3.5}$ & $76.73{\scriptstyle \pm1.0}$ \\
% BERT-uncased & $0.2069{\scriptstyle \pm0.0027}$ & $86.08{\scriptstyle \pm0.6}$ & $87.09{\scriptstyle \pm0.6}$ & $85.69{\scriptstyle \pm0.3}$ & $70.89{\scriptstyle \pm1.1}$ & $76.70{\scriptstyle \pm0.7}$ \\
% FinancialBERT & $0.2500{\scriptstyle \pm0.0062}$ & $84.96{\scriptstyle \pm0.6}$ & $83.53{\scriptstyle \pm0.9}$ & $85.90{\scriptstyle \pm0.3}$ & $67.52{\scriptstyle \pm1.6}$ & $75.39{\scriptstyle \pm1.1}$ \\
% Y-FinBERT & $0.2275{\scriptstyle \pm0.0061}$ & $85.62{\scriptstyle \pm1.2}$ & $86.55{\scriptstyle \pm0.6}$ & $85.66{\scriptstyle \pm0.6}$ & $65.45{\scriptstyle \pm2.5}$ & $74.75{\scriptstyle \pm1.3}$ \\
% P-FinBERT & $\mathbf{0.2054}{\scriptstyle \pm0.0057}$ & $\mathbf{86.61}{\scriptstyle \pm0.5}$ & $\mathbf{87.67}{\scriptstyle \pm0.2}$ & $\mathbf{85.74}{\scriptstyle \pm0.5}$ & $\mathbf{72.66}{\scriptstyle \pm3.3}$ & $\mathbf{77.12}{\scriptstyle \pm0.8}$ \\
% \hline
% \end{tabular}
% \caption{\label{tab:stl_baselines}
% Performance of various STL baselines. We use the best-performing one, P-FinBERT as the backbone of our MTL model.
% }
% \vspace{-0.7em}
% \end{table*}
\myparagraph{Pre-trained Model Selection \& STL Baselines:} Existing financial pre-trained models \citep{araci2019finbert,yang2020finbert,liu2021finbert,financialbert2022} are usually compared on the Financial PhraseBank dataset \citep{financialphrasebank2013malo}. Such comparison is suboptimal because (1) Financial PhraseBank sentiment analysis has no official test set. Existing work separates test sets on their own, making the scores less comparable across different work; and (2) the models are not compared on benchmarks other than sentiment analysis. Therefore, we compare financial pre-trained models on all FinDATA tasks to select the best one.

STL results on all publicly available financial pre-trained models (P-FinBERT \citep{araci2019finbert}, Y-FinBERT \citep{yang2020finbert}, and FinancialBERT \citep{financialbert2022}) and BERT \cite{devlin-etal-2019-bert} are presented in the first half of \cref{tab:mtl}. P-FinBERT \citep{araci2019finbert} outperforms other pre-trained models. Therefore, we use P-FinBERT in all subsequent experiments.

% \begin{figure}[t]
%     \centering
%     \includegraphics[width=\columnwidth]{pictures/spal_finbert.png}
%     \caption{Visualization of a SPAL-FinBERT layer.}
%     \label{fig:spallayer}
%     \vspace{-1.5em}
% \end{figure}

%When exploring the importance of \textbf{skill diversity} and \textbf{task-relatedness}, we use FinBERT \citep{araci2019finbert} as the shared encoder. When exploring \textbf{aggregation size versus model capacity}, we use SPAL-FinBERT (detailed introduction in \cref{sec:factor_3}) instead to enable tuning of the model capacity.
\section{Analysis}
In this section, we analyze the hypotheses about when aggregating multiple skills with MTL works.
%\subsection{Does Skill Diversity Benefit MTL?} 
\subsection{
H1: 
% When There's 
Skill Diversity} \label{sec:skill_diversity}
\begin{table*}[t]
\centering
%\small
%\centering
\resizebox{\textwidth}{!}{
\begin{tabular}{lllcclcclcccc}
\hline
\multirow{2}{*}{Method} & \multicolumn{1}{c}{\multirow{2}{*}{\begin{tabular}[l]{@{}l@{}}STL Model or\\ MTL Subset\end{tabular}}} &  & \multicolumn{2}{c}{Sentiment}                                   &  & \multicolumn{2}{c}{Number}                                  &  & -                            &                      & -                            &                      \\ \cline{4-5} \cline{7-8} \cline{10-10} \cline{12-12}
                        & \multicolumn{1}{c}{}                                                                                     &  & TSA$\downarrow$                              & SC                           &  & NC                           & NAD                          &  & FSRL                         &                      & CD                           &                      \\ \hline
\multirow{5}{*}{STL}    & BERT-cased                                                                                                &  & $0.2320{\scriptstyle \pm0.0082}$ & $86.57{\scriptstyle \pm0.8}$ &  & $86.19{\scriptstyle \pm1.1}$ & $85.43{\scriptstyle \pm0.8}$ &  & $71.30{\scriptstyle \pm3.5}$ & \multicolumn{1}{l}{} & $76.73{\scriptstyle \pm1.0}$ & \multicolumn{1}{l}{} \\
                        & BERT-uncased                                                                                              &  & $0.2069{\scriptstyle \pm0.0027}$ & $86.08{\scriptstyle \pm0.6}$ &  & $87.09{\scriptstyle \pm0.6}$ & $85.69{\scriptstyle \pm0.3}$ &  & $70.89{\scriptstyle \pm1.1}$ & \multicolumn{1}{l}{} & $76.70{\scriptstyle \pm0.7}$ & \multicolumn{1}{l}{} \\
                        & FinancialBERT                                                                                             &  & $0.2500{\scriptstyle \pm0.0062}$ & $84.96{\scriptstyle \pm0.6}$ &  & $83.53{\scriptstyle \pm0.9}$ & $85.90{\scriptstyle \pm0.3}$ &  & $67.52{\scriptstyle \pm1.6}$ & \multicolumn{1}{l}{} & $75.59{\scriptstyle \pm1.1}$ & \multicolumn{1}{l}{} \\
                        & Y-FinBERT                                                                                                 &  & $0.2275{\scriptstyle \pm0.0061}$ & $85.62{\scriptstyle \pm1.2}$ &  & $86.55{\scriptstyle \pm0.6}$ & $85.66{\scriptstyle \pm0.6}$ &  & $65.45{\scriptstyle \pm2.5}$ & \multicolumn{1}{l}{} & $74.75{\scriptstyle \pm1.3}$ & \multicolumn{1}{l}{} \\
                        & P-FinBERT                                                                                                 &  & $\dashuline{\textbf{0.2054}}{\scriptstyle \pm0.0057}$ & $\dashuline{86.61}{\scriptstyle \pm0.5}$ &  & $\dashuline{87.67}{\scriptstyle \pm0.6}$ & $\dashuline{85.74}{\scriptstyle \pm0.5}$ &  & $\dashuline{\textbf{72.66}}{\scriptstyle \pm3.3}$ &                      & $\dashuline{77.12}{\scriptstyle \pm0.8}$ &                      \\ \hline
\multirow{7}{*}{MTL}    & Full FinDATA                                                                                              &  & $0.2151{\scriptstyle \pm0.0089}$ & $\underline{\textbf{87.06}}{\scriptstyle \pm1.1}$ &  & $87.51{\scriptstyle \pm0.7}$ & $\underline{\textbf{86.52}}{\scriptstyle \pm0.4}$ &  & $69.88{\scriptstyle \pm1.5}$ &                      & $77.80{\scriptstyle \pm0.8}$ &                      \\
                        & w/o FSRL                                                                                                  &  & $0.2156{\scriptstyle \pm0.0099}$ & $85.91{\scriptstyle \pm1.4}$ &  & $87.41{\scriptstyle \pm0.8}$ & $86.11{\scriptstyle \pm0.6}$ &  & -                            &                      & $76.53{\scriptstyle \pm0.8}$ &                      \\
                        & w/o CD                                                                                                    &  & $\underline{0.2077}{\scriptstyle \pm0.0032}$ & $86.36{\scriptstyle \pm0.8}$ &  & $87.49{\scriptstyle \pm0.4}$ & $85.63{\scriptstyle \pm0.6}$ &  & $\underline{71.32}{\scriptstyle \pm1.8}$ &                      & -                            &                      \\
                        & w/o Sentiment                                                                                             &  & -                                & -                            &  & $\underline{\textbf{87.79}}{\scriptstyle \pm1.0}$ & $86.49{\scriptstyle \pm0.5}$ &  & $70.60{\scriptstyle \pm3.0}$ &                      & $\underline{\textbf{78.40}}{\scriptstyle \pm1.0}$ &                      \\
                        & w/o Number                                                                                                &  & $0.2083{\scriptstyle \pm0.0046}$ & $86.49{\scriptstyle \pm1.2}$ &  & -                            & -                            &  & $71.08{\scriptstyle \pm2.6}$ & \multicolumn{1}{l}{} & $78.26{\scriptstyle \pm1.0}$ & \multicolumn{1}{l}{} \\
                        & Only Sentiment                                                                                            &  & $0.2159{\scriptstyle \pm0.0120}$ & $86.69{\scriptstyle \pm1.1}$ &  & -                            & -                            &  & -                            & \multicolumn{1}{l}{} & -                            & \multicolumn{1}{l}{} \\
                        & Only Number                                                                                               &  & -                                & -                            &  & $87.25{\scriptstyle \pm1.0}$ & $85.70{\scriptstyle \pm0.5}$ &  & -                            &                      & -                            &                      \\ \hline
\end{tabular}
}
\caption{\label{tab:mtl}
The first half (STL in the method column) shows the performance of various STL baselines. We use the best-performing one, P-FinBERT, as the backbone of our MTL model. The second half (MTL in the method column) shows the MTL results on FinDATA and its subsets. The \textbf{bold} numbers denote the best scores obtained in all settings. The \dashuline{dashed underline} numbers denote the best STL baselines. The \underline{underlined} numbers denote the best scores obtained with MTL. 
}
%\vspace{-0.5em}
\end{table*}
To verify the hypothesis that skill diversity benefits MTL, we compare the MTL results on full FinDATA and its subsets that ablate one skill or focus on one skill. Specifically, ablating a skill results in four subsets: w/o financial semantic role labeling, w/o causality detection, w/o sentiment analysis, and w/o number understanding. Focusing on a single skill results in two subsets: only sentiment analysis and only number understanding. We use FinBERT \citep{araci2019finbert} as the shared encoder. The results are shown in \cref{tab:mtl}. It can be observed that (1) skill diversity benefits MTL: the best MTL scores of all tasks are obtained by mixing several different skills while concentrating on sentiment/number understanding skills (w/o Sentiment and w/o Number) leads to a performance drop on corresponding tasks; and (2) ablating FSRL decreases the performance of all other tasks, illustrating that FSRL positively transfers to all other skills. %This is surprising as the form of FSRL (token classification) is different from many others, and the dataset is limited. 
Therefore, positive transfers can happen between different skills. \textbf{Including skills other than the target skill in MTL is a potential way to benefit target performance.} %(3) Ablating sentiment analysis skills results in the best scores on NC and CD. Similarly, ablating number understanding skills improves the TSA, FSRL, and CD performance. Therefore, the degree of skill diversity needs to be \textbf{appropriate} as a negative transfer happens among certain skills.

%\subsection{Does task-relatedness matters?} 
\subsection{H2: 
Task Relatedness
% When Tasks are Well Related
} \label{sec:task_relatedness}
Similar to FinDATA, GLUE also aggregates multiple NLU skills. However, GLUE MTL usually leads to a performance drop on most tasks (according to \citet{stickland2019pal} only RTE is improved; and according to \citet{mueller2022t2t}, 3 out of 9 tasks are improved) while FinDATA MTL increases the scores of 4 out of 6 included tasks. Therefore, we hypothesize that FinDATA tasks are more closely related than GLUE tasks though they all cover diverse skills. We measure the relatedness among FinDATA tasks qualitatively and quantitatively:

\myparagraph{Qualitative Analysis:} Many tasks relate to each other explicitly: (1) SC and TSA: though they have different annotations, both of them predict financial sentiment. %(2)  NC and NA: in NA, monetary numbers are usually attached to the target stock, and NC helps to understand monetary numbers. 
(2) FSRL and NC: ``date'' is one of the classes in NC, while FSRL helps to understand the semantic role of time numbers. These explicit transfers can be probed by different output headers of an MTL system: for an input sentence, the MTL system outputs predictions corresponding to different tasks, where the non-target headers' predictions may interpret the target prediction \citep{geva-etal-2021-whats}. In \cref{sec:transfer_example}, we illustrate these explicit transfers by listing examples of the prediction header's outputs. 

\begin{figure*}[t]
     \centering
     \begin{subfigure}[t]{0.47\textwidth}
         \centering
         \includegraphics[width=\textwidth]{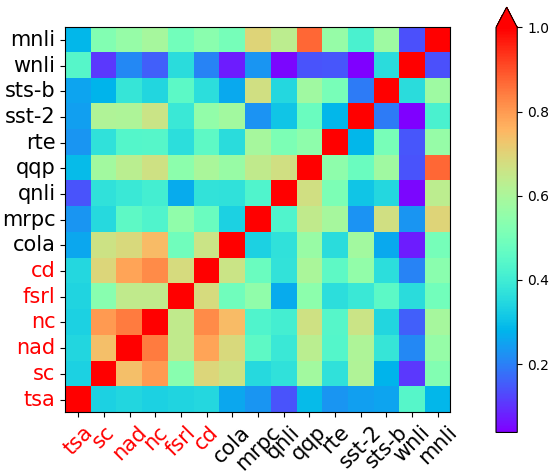}
         \vspace{-1.7em}
         \caption{TaskEmb similarity heatmap.}
         \label{fig:tasksim}
     \end{subfigure}
     \hfill
     \begin{subfigure}[t]{0.47\textwidth}
         \centering
         \includegraphics[width=\textwidth]{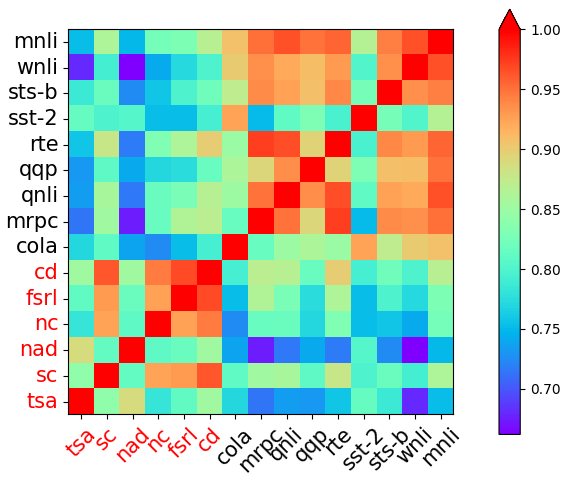}
         \vspace{-1.7em}
         \caption{TextEmb similarity heatmap}
         \label{fig:textsim}
     \end{subfigure}
        \vspace{-0.5em}
        \caption{Heatmaps of cosine similarity between TaskEmbs and TextEmbs. FinDATA tasks are \textcolor{red}{highlighted in red} on both axes.}
        \label{fig:heatmaps}
        %\vspace{-1.5em}
\end{figure*}

\myparagraph{Quantitative Analysis:} \citet{vu-etal-2020-exploring} propose task and text embedding to measure the similarity between task objectives and texts. This embedding algorithm facilitates high-level knowledge sharing in the MTL architecture proposed by \citet{karimi-mahabadi-etal-2021-parameter}, which achieves superior performance. Therefore, we use these metrics to quantify the relatedness among the tasks aggregated in our MTL systems. We follow \citeposs{vu-etal-2020-exploring} calculation setting, except that we use FinBERT instead of BERT: we first calculate task and text embeddings of FinDATA and GLUE tasks. Then we compute the cosine similarity scores among embeddings. 

\cref{fig:tasksim} shows the heatmap of task embedding similarity scores, indicating that \textbf{FinDATA tasks are more closely clustered than GLUE tasks}, illustrating why FinDATA MTL leads to more improvements than GLUE MTL. Another observation is that \textbf{TSA has the lowest similarity scores with other FinDATA tasks}, which possibly explains why it is not improved by MTL in \cref{tab:mtl}. \cref{fig:textsim} presents the heatmap of text embedding similarity, where financial and general data are well separated with high in-domain similarity. 

However, the similarity scores are symmetric metrics and thus fail to explain some asymmetric transfers (which is also observed in previous work \citep{geva-etal-2021-whats}). For example, FSRL has a moderate level of text and task similarity to other tasks, but its performance is not enhanced by MTL while it boosts the performance of others. A possible explanation is that financial semantic understanding skill (provided by FSRL) is a necessary ability for other FinDATA tasks, but the skills covered by other tasks are not necessary for FSRL. Therefore, the joint training does not benefit FSRL.

We further analyze whether gradient similarities interpret task-relatedness and MTL transferability since many previous works attribute the negative transfer among aggregated tasks to gradient conflicts \citep{ruder2017mtl,gradient2020yu,wang2021gradient,mueller2022t2t}. However, our findings in \cref{sec:gradient_analysis} show that gradient conflicts/similarities are not good measurements.

In conclusion, the degree of task-relatedness serves as a significant predictor of the MTL outcome, and can be roughly measured through quantitative and qualitative means. To better explain asymmetric transfer and analyze the inter-task relations in a finer grain, it is essential to develop asymmetric measurements. We reserve that exploration for future work.
%\subsection{Should Aggregation Size Matches Shared Capacity?} 
\subsection{H3: Matched 
% When 
Aggregation Size
% Matches 
with
Shared Capacity} \label{sec:model_capacity}
\begin{figure}[t]
    \centering
    \includegraphics[width=\columnwidth]{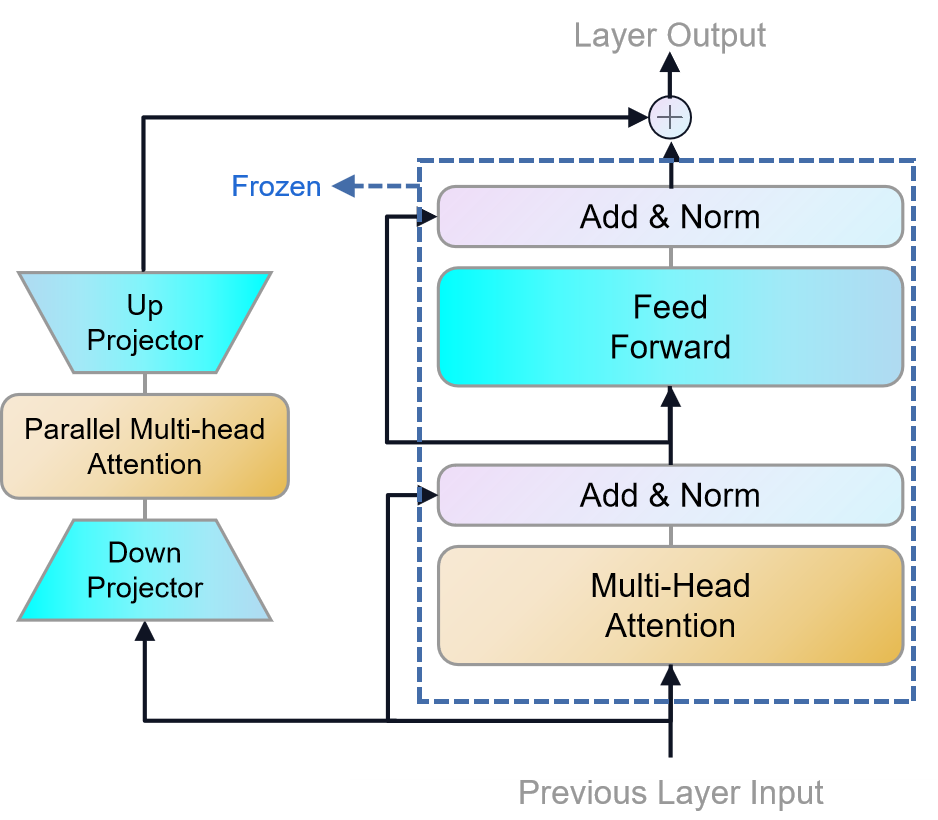}
    \caption{Visualization of a SPAL-FinBERT layer.}
    %\vspace{-1.5em}
    \label{fig:spallayer}
\end{figure}
\begin{figure*}[t]
    \centering
    \includegraphics[width=\textwidth]{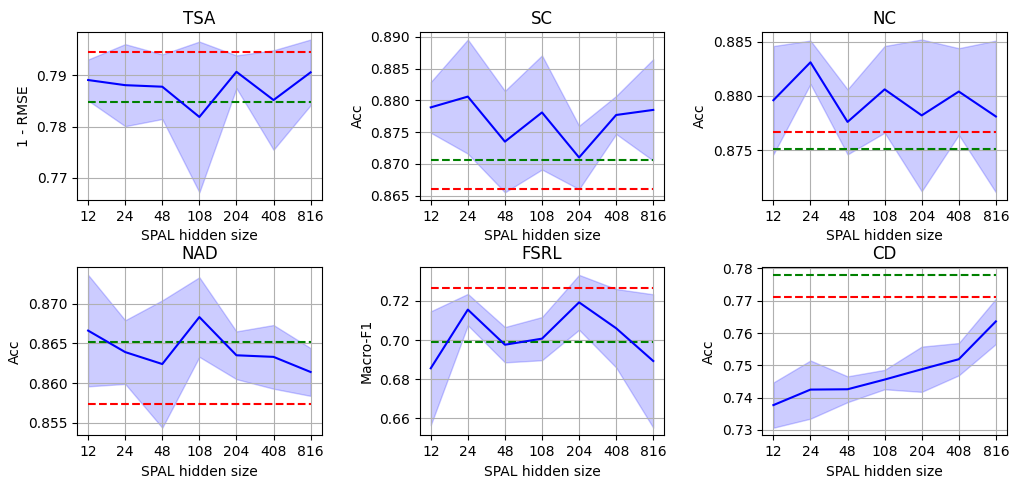}
    \caption{FinDATA MTL results with different shared capacities. The solid blue lines ``\textcolor{blue}{---}'' denote the average SPAL-FinBERT MTL results of 5 random seeds and their standard deviations. The dashed red lines ``\textcolor{red}{- -}'' denote the STL results. The dashed green lines ``\textcolor{green}{- -}'' denote the vanilla FinBERT MTL results.}
    %\vspace{-0.5em}
    \label{fig:aggregation_vs_capacity}
\end{figure*}
We hypothesize that having too many tasks sharing limited model capacity might cause interference among tasks and result in poor MTL performance. Therefore, given a fixed pre-trained model, the task aggregation size should be appropriate for the shared capacity to achieve the best MTL practice.

\cref{sec:skill_diversity} shows that the task combination significantly influences the MTL performance. Altering the task aggregation may introduce unwanted positive or negative transfers. Therefore, to verify this hypothesis, we control the task aggregation (stick to FinDATA instead of adding other tasks) and reduce the shared capacity to simulate the scenario where task aggregation may exhaust the shared capacity.

To enable altering shared capacity, we propose SPAL-FinBERT, an architecture that both leverages a pre-trained model and has tunable shared capacity. \cref{fig:spallayer} illustrates the architecture. The FinBERT layers are frozen while the parallel attention layers (PALs, \citealp{stickland2019pal}) are trainable. Different from original task-specific PALs, ours are shared across different tasks. Thus, we call them shared PALs (SPALs). The design is similar to Adapters \citep{houlsby2019parameterefficient}: both consists of light-weighted trainable structures and a frozen pre-trained model. We choose PAL as the shared trainable structure because it has a more complicated structure than an adapter which might benefit multi-task knowledge sharing (Adapters are usually for STL). We can easily change the shared capacity by setting the SPAL hidden size to any multiple of 12 (the number of self-attention heads).

\begin{table*}[t]
\centering
\begin{tabular}{cccccc}
\hline
 TSA$\downarrow$               & SC                           & NC                           & NAD                          & FSRL                         & CD                                    \\ \hline
 $0.2109{\scriptstyle \pm0.0040}$ & $87.89{\scriptstyle \pm0.4}$ & $87.97{\scriptstyle \pm0.5}$ & $86.66{\scriptstyle \pm0.7}$ & $68.56{\scriptstyle \pm2.9}$ & $\adjustbox{margin=2pt,bgcolor=green!0}{73.77}{\scriptstyle \pm0.7}$          \\
      -                                & $87.31{\scriptstyle \pm0.5}$ & $88.09{\scriptstyle \pm0.6}$ & $86.30{\scriptstyle \pm0.3}$ & $68.35{\scriptstyle \pm3.4}$ & $\adjustbox{margin=2pt,bgcolor=green!0}{74.31}{\scriptstyle \pm0.5}$          \\
      -                                & -                            & $87.73{\scriptstyle \pm0.3}$ & $85.80{\scriptstyle \pm0.4}$ & $70.66{\scriptstyle \pm1.8}$ & $\adjustbox{margin=2pt,bgcolor=green!0}{73.61}{\scriptstyle \pm0.8}$          \\
      -                                & -                            & -                            & $85.99{\scriptstyle \pm0.5}$ & $69.80{\scriptstyle \pm2.7}$ & $\adjustbox{margin=2pt,bgcolor=green!0}{74.65}{\scriptstyle \pm0.6}$          \\
      -                                & -                            & -                            & -                            & $67.27{\scriptstyle \pm0.8}$ & $\adjustbox{margin=2pt,bgcolor=green!0}{\textbf{74.80}}{\scriptstyle \pm0.5}$ \\
      -                                & -                            & -                            & -                            & -                            & $73.57{\scriptstyle \pm0.8}$          \\ \hline
\end{tabular}
\caption{\label{tab:mtl_limited_capacity}
MTL results on SPAL-FinBERT with minimal shared capacity (SPAL hidden size = 12). Gradually decreasing the number of aggregated tasks improves CD performance in general.
}
%\vspace{-0.5em}
\end{table*}
We run FinDATA MTL with SPAL hidden size from 12 to 816. The smallest and the largest trainable shared capacity are roughly 228K\footnote{SPALs' parameter number is calculated by \#layers$\times$(self-attention layer$+$projection layer), e.g., $12\times(204\times204\times4 + 204\times768\times2)\approx5.8$M when hidden size is 204.} ($0.2\%$ of FinBERT parameters) and 47M ($42.7\%$ of FinBERT parameters). The results are shown in \cref{fig:aggregation_vs_capacity}. We surprisingly find that the aggregated tasks are not equally sensitive to the change of shared capacity: negative transfer towards CD grows while the shared capacity becomes limited. However, Some tasks are not significantly restricted by the limited shared capacity: SC and NC even achieve the best scores with relatively small shared capacity.

To verify that aggregating too many tasks in limited capacity overwhelms CD, we gradually ablating tasks from the MTL system with minimal shared capacity. \cref{tab:mtl_limited_capacity} presents the results. The CD performance gradually improves when we decrease the aggregation size (although the task combination can be a confounder for the CD performance). The highest score is achieved when only aggregating two tasks. 

Therefore, to achieve better MTL practice, \textbf{the aggregation size should be appropriate for the shared capacity to avoid overwhelming tasks like CD}. These tasks are sensitive to capacity sharing. Including too many auxiliary objectives might exhaust the shared capacity, distracting the MTL system from these tasks. Other tasks (e.g., SC and NC) might be more tolerant for capacity sharing, thus allowing larger-scale MTL auxiliary training. 

\subsection{Efficiency of SPAL-FinBERT}
\begin{figure}[t]
    \centering
    \includegraphics[width=\columnwidth]{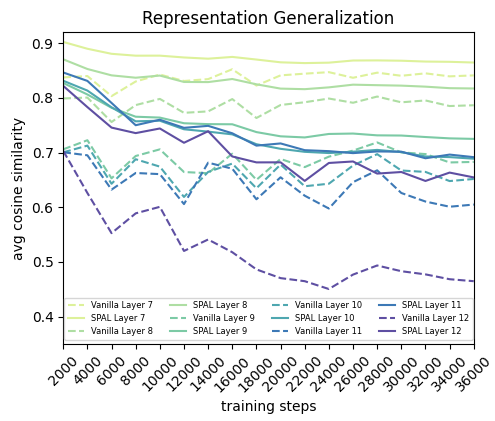}
    \caption{Representation generalization of SPAL-FinBERT and vanilla FinBERT at different training steps, measured with a SPAL hidden size of 204 (PAL setting recommended by \citet{stickland2019pal}).}
    %\vspace{-0.7em}
    \label{fig:generalization}
\end{figure}
Another observation of \cref{fig:aggregation_vs_capacity} is that SPAL-FinBERT outperforms vanilla FinBERT with much fewer trainable parameters. In the most impressive case, SPAL-FinBERT outperforms vanilla FinBERT on four tasks with $99.8\%$ fewer trainable parameters (when the SPAL hidden size is 12). One possible reason behind our model's performance is that the frozen FinBERT provides a strong regularization and thus reduces the representation variance of each layer. Such generalized representations are more likely to be favored by different tasks and thus benefit MTL \citep{ruder2017mtl}. To verify this explanation, we compare the representation generalizations of SPAL-FinBERT and FinBERT MTL systems.

\myparagraph{Representation generalization:} Intuitively, representation generalization measures how similar an MTL system represents data of different tasks. We first compute the representations for all tasks, models, and layers, following the formula:
\begin{equation}
    \mathbf{R_{\mathit{l},\mathcal{M}}^{\mathit{t}}} = \frac{1}{|D_{\mathit{t}}|} \sum_{(\mathbf{\mathit{x}}_{\mathit{t}},y_{\mathit{t}}) \in D_{\mathit{t}}} \mathcal{M}_{\mathit{l}}(\mathbf{\mathit{x}}_{\mathit{t}})
    %\vspace{-0.7em}
\end{equation}
where $\mathbf{R_{\mathit{l},\mathcal{M}}^{\mathit{t}}}$ denotes task $t$'s representation generated by layer $\mathit{l}$ of MTL model $\mathcal{M}$; $D_{\mathit{t}}$ denotes the dataset of task $t$; and $(\mathbf{\mathit{x}}_{\mathit{t}},y_{\mathit{t}})$ denotes the data points. Then, we compute the cosine similarity score between all task representation pairs $(\mathbf{R_{\mathit{l},\mathcal{M}}^{\mathit{t_1}}}, \mathbf{R_{\mathit{l},\mathcal{M}}^{\mathit{t}_2}})$, averaging the similarity scores to measure the representation generalization of model $\mathcal{M}$ layer $\mathit{l}$:
\begin{equation}
    \mathbf{G_{\mathit{l},\mathcal{M}}}=\frac{1}{C^{2}_{\mathbf{|T|}}} \sum_{\mathit{t_1},\mathit{t_2} \in \mathbf{T}} \mathrm{cossim}(\mathbf{R_{\mathit{l},\mathcal{M}}^{\mathit{t_1}}}, \mathbf{R_{\mathit{l},\mathcal{M}}^{\mathit{t}_2}})
    %\vspace{-0.7em}
\end{equation}
where $C$ denotes combination, $\mathbf{T}$ denotes the task collection, and $\mathrm{cossim}$ denotes cosine similarity. \cref{fig:generalization} shows the representation generalization for two MTL systems at different training steps. For simplicity, only higher layers' results (layer $7$ to $12$) are presented as they are modified more by fine-tuning \citep{zhou-srikumar-2022-closer} and related more to the output. It can be observed that SPAL-FinBERT generates more generalized representations than FinBERT in all shown layers (especially for the highest ones).

Another observation is that representation generalization decreases when the training step increases. One possible explanation for this downward trend is that the MTL system is trying to learn task-specific knowledge (especially in higher layers) as multi-task fine-tuning continues. %SPAL-FinBERT effectively regularizes this trend, thus learning a more generalized representation that promotes positive transfer.

In \cref{appendix:freeze}, we further use an ablation experiment and a probing experiment to show the contribution of the frozen FinBERT and the necessity of freezing.

\section{Discussion}
\myparagraph{Suggestions for MTL practice:} Based on the results of our case study, we recommend the following practices for future MTL: (1) aggregate not only the target skill but also other related skills; (2) select tasks for aggregation based on both their qualitative and quantitative relatedness to the target task; and (3) check if the target task is sensitive to capacity sharing, excluding redundant (e.g., distantly related) tasks to avoid distracting the target task.

% \myparagraph{task-relatedness does not approximately indicate MTL transferability}: \cref{fig:heatmaps} shows that the FinDATA tasks are more closely related than GLUE tasks, explaining why positive transfer happens more often on FinDATA MTL. However, these embedding similarity scores do not fully explain all the observations in \cref{tab:mtl}. For example, two financial number-related tasks (NAD and NC) have very similar task embeddings but do not positively transfer to each other when only these tasks are aggregated. Besides, FSRL is less similar to other tasks but shows good transferability to other tasks in MTL. This transferability is also asymmetric, as other FinDATA tasks do not benefit FSRL performance. In \cref{sec:gradient_analysis}, We further analyze whether gradient similarities interpret task-relatedness and MTL transferability since many previous works attribute the negative transfer among aggregated tasks to gradient conflicts \citep{ruder2017mtl,gradient2020yu,wang2021gradient}. However, gradient conflict/similarity also fails to explain the MTL phenomena.

% Therefore, future work may explore if there is better indicators for the transferability within a task aggregation, which will be very helpful for identifying the optimal task grouping for a target task.

\myparagraph{Aggregating multiple skills with MTL is a potential way for better Financial NLP practice:} Financial NLP tasks are more complicated than those in the general domain, and many of them suffer from a lack of data. Obtaining new Financial NLP data is expensive since such annotation usually requires domain expertise. Our results show that aggregating Financial NLP tasks using MTL can be a practical and relatively cheap way to improve their performance: SC, NC, NAD, and CD are improved by up to $1.45$, $0.64$, $1.09$, and $0.68$ percentage points accordingly through MTL auxiliary training (contributed by different MTL systems). In \cref{sec:generalization}, we also show that MTL pre-training with Financial NLP tasks can improve the model's generalizability to unseen tasks. Therefore, future research and practice in Financial NLP may consider MTL as a potential way to achieve better performance.
\myparagraph{Other possible questions:} We address some other possible questions that might be of interest to our readers in \cref{sec:PQA}.

\section{Conclusion}
In this work, we conduct a case study on Financial NLP to analyze when aggregating multiple skills with MTL works from a perspective of task relations and skills to be included. We propose a parameter-efficient MTL architecture SPAL-FinBERT. Our empirical analyses point out potential directions to improve task aggregation for future MTL practice: (1) considering diversified non-target skills that might be supportive; (2) filtering tasks with their relatedness; and (3) caring whether capacity sharing overwhelms the target task. %\mrinmaya{What is this direction? Can we restate in a line or two?}
We also show that aggregating resources through MTL can be a cheap and efficient way to improve Financial NLP performance.

\section*{Limitations}
Firstly, the transferability between different tasks within an MTL system is not well measured in current work. We also find such transferability is asymmetric and thus hard to quantify using symmetric measurements such as cosine similarity between task embeddings or gradients: for example, TSA positively transfers to SC, but SC negatively transfers to TSA (see the ``Only Sentiment'' row in \cref{tab:mtl}); FSRL positively transfer to all other tasks, but other tasks negatively affect FSRL. Future work may consider exploring better indicators that address the asymmetry of task transferability (e.g., similar to inter-task affinity scores \citep{fifty2021efficiently} in the CV domain).

Secondly, some of the conclusions drawn from our case study only point in a vague direction for future MTL practice. For example, we find that some tasks are more sensitive to capacity sharing in \cref{sec:model_capacity}. Therefore, aggregating an excessive number of tasks with those tasks might overwhelm them. However, it is hard to determine exactly each task's sensitivity to capacity sharing and the optimal number of aggregated tasks without some trials on different task combinations. Future work may explore why some tasks are easily overwhelmed by capacity sharing and propose methods to identify them.

Thirdly, in this work, we analyze the influence of multiple factors on MTL performance. However, the factors are usually entangled and confound each other. For example, we decrease the number of tasks aggregated with CD to show that too large aggregation overwhelms CD in a limited shared capacity. But the tasks included (a confounder for MTL performance) are also changed. Future work may conduct rigorous causal analyses, exploring how much each factor affects MTL performance.

\section*{Ethical Considerations}
% Scientific work published at ACL 2023 must comply with the ACL Ethics Policy.\footnote{\url{https://www.aclweb.org/portal/content/acl-code-ethics}} We encourage all authors to include an explicit ethics statement on the broader impact of the work, or other ethical considerations after the conclusion but before the references. The ethics statement will not count toward the page limit (8 pages for long, 4 pages for short papers).
\myparagraph{Data Privacy and Bias}: All datasets used in this research are published in previous studies and publicly available: datasets for TSA, SC, FSRL, CD, and Numeracy-600K can be downloaded from the internet, while datasets for NC, NAD, and StockSen require signing corresponding agreements and requesting from the authors.

Licenses: TSA is under Apache License 2.0; SC is under CC BY-NC-SA 3.0; CD and StockSen are under CC BY 4.0; and Numercay-600K, NC, and NAD are under CC BY-NC-SA 4.0. The license of FSRL data is not explicitly specified, but the author allows data usage with a proper citation in their GitHub repository.

Most of the datasets are widely used in the Financial NLP domain (e.g., shared tasks). We also manually checked for offensive content in the data. There is no data bias against certain demographics with respect to these datasets.

\myparagraph{Reproducibility}:
\ifarxiv
We make all of our code public on GitHub.
\else
We will open-source our codes on GitHub upon acceptance.
\fi
For data, we include links to request NC, NAD, and StockSen, and provide data splits for TSA, SC, FSRL, CD, and Numeracy-600K. We also provide detailed instructions to reproduce all the experiment results on GitHub.

\myparagraph{Potential Use}: The potential use of this study is to improve future practice in MTL and the Financial NLP domain.

\ifarxiv
\section*{Author Contributions}
\myparagraph{Jingwei Ni} designed the project and the storyline, and conducted the MTL analyses and the survey in Financial NLP. \\

\myparagraph{Zhijing Jin} helped design the storyline and provided essential suggestions on what experiments and analyses are important. \\

\myparagraph{Qian Wang} contributed to the financial background of the storyline, collected the first version of FinDATA, and gave insights on what skills are important from a financial perspective. \\

\myparagraph{Mrinmaya Sachan} and \textbf{Markus Leippold} guided the project and substantially contributed to the storyline and experiment design. \\

\myparagraph{Everyone} contributed to writing the paper.

\section*{Acknowledgements}
We sincerely thank the authors of \citet{chen2020overview}, \citet{chen2022overview}, and \citet{xing-etal-2020-financial} for granting us access to their proposed datasets for research.

\fi

\bibliography{anthology,custom}
\bibliographystyle{acl_natbib}

\appendix
\section{Possible Questions and Answers} \label{sec:PQA}

\subsection{Are H1 and H2 proposed in our work conflict goals?}
H1 encourages aggregating diverse skills for better MTL practice, while H2 suggests that the relatedness among tasks is also important, which seems contrary to H1. However, these goals do not conflict with each other because: (1) skill diversity does not imply distant inter-task relationships and vice versa (e.g., NC is well related to SC and CD in \cref{fig:tasksim} though they correspond to different skills). (2) It is possible to achieve skill diversity and good task-relatedness simultaneously: real-world MTL practice can first consider the target skill and other skills that might be supportive or in the same domain. Then select the tasks that are (qualitatively or quantitatively) closely related to the target task to achieve better MTL performance.

\subsection{Our work mainly addresses Financial NLP. Are the conclusions generalizable to other domains?}
Although we only provide analyses in Financial NLP, the heuristics for MTL practice are generic for other domains. For H1, non-target skills in the same domain are potentially helpful as various skills are based on similar data. For H2, the qualitative and quantitative analyses for task-relatedness are domain agnostic, meaning that we can select the most related tasks from those with diversified skills. For H3, continuously increasing the aggregation size will finally reach a threshold that overwhelms some tasks if the shared capacity is fixed.

\section{GPT-3 Prompts} \label{appendix:gpt3}
In \cref{tab:gpt-3} we present the GPT-3 zero-shot and few-shot performance on two Financial NLP tasks. We use the official API provided by OpenAI\footnote{https://openai.com/api/} to access GPT-3. We choose the GPT-3 checkpoint Davinci-003 to conduct the experiments (completion mode, max token $5$, temperature $0$). The example prompts we use for TSA and SC are illustrated in \cref{tab:prompts}.

\begin{table*}[t]
\centering
\resizebox{\textwidth}{!}{
\begin{tabular}{lll}
\hline
Task & Example Prompts                                                                                                                                                                                                                                               & Label \\ \hline
TSA zero-shot  & \begin{tabular}[c]{@{}l@{}}"Financial sentiment refers to the prevailing emotions and opinions of investors and traders towards\\ financial markets or specific investments, which can be positive or negative and influence buying or \\selling decisions.\\Decide a financial sentiment score between -1 and 1 about the target company in a news headline.\\ \\ Headline: Ashtead to buy back shares, full-year profit beats estimates.\\ Company: Ashtead\\ Sentiment score:"\end{tabular} & 0.588           \\ \hline
SC zero-shot  & \begin{tabular}[c]{@{}l@{}}"Financial sentiment refers to the prevailing emotions and opinions of investors and traders towards\\ financial markets or specific investments, which can be positive or negative and influence buying or \\selling decisions.\\Decide whether a news headline's financial sentiment is positive, neutral, or negative.\\ \\ Headline: An Android app will be coming soon.\\ Sentiment:"\end{tabular} & Neutral        \\ \hline
TSA few-shot  & \begin{tabular}[c]{@{}l@{}}"Financial sentiment refers to the prevailing emotions and opinions of investors and traders towards\\ financial markets or specific investments, which can be positive or negative and influence buying or\\ selling decisions.\\Given a few examples, decide a financial sentiment score between -1 and 1 about the target \\ company in a news headline.\\ \\ Headline: Brazil Vale says will appeal ruling to block assets for dam burst \\ Company: Vale\\ Sentiment score: -0.131\\ \\ Headline: Sainsbury's share price: Grocer launches click-and-collect \\ Company: Sainsbury's\\ Sentiment score: 0.021\\ \\ Headline: Rolls-Royce Wins \$9.2 Billion Order From Emirates Airline \\ Company: Rolls-Royce\\ Sentiment score: 0.777\\ \\ Headline: Ashtead to buy back shares, full-year profit beats estimates.\\ Company: Ashtead\\ Sentiment score:"\end{tabular} & 0.588           \\ \hline
SC few-shot  & \begin{tabular}[c]{@{}l@{}}"Financial sentiment refers to the prevailing emotions and opinions of investors and traders towards\\ financial markets or specific investments, which can be positive or negative and influence buying or\\ selling decisions.\\Given a few examples, decide whether a news headline's financial sentiment is positive, neutral, \\or negative.\\ \\ Headline: The business to be divested generates consolidated net sales of EUR 60 million \\ annually and currently has some 640 employees.\\ Sentiment: neutral\\ \\ Headline: Cargo volume increased by approximately 5 \% .\\ Sentiment: positive\\ \\ Headline: Operating loss increased to EUR 17mn from a loss of EUR 10.8 mn in 2005.\\ Sentiment: negative\\ \\ Headline: An Android app will be coming soon.\\ Sentiment:"\end{tabular} & Neutral        \\ \hline
\end{tabular}
}
\caption{\label{tab:prompts} Examples of prompts used by us for running target-based sentiment analysis (TSA) and sentiment classification (SC) on GPT-3 (text-davinci-003).}
\end{table*}

\section{Experimental Details} \label{sec:experimental_details}
\myparagraph{MTL Batching Scheme}: During MTL, we first randomly batchify training data of all tasks. Then, we randomly mix the mini-batches and pass them to the MTL data loader. This method is equivalent to the temperature-based batch sampling scheme of \citet{karimi-mahabadi-etal-2021-parameter} where temperature $T=1$ (i.e., each task is sampled proportional to its data size). We choose $T=1$ as FinDATA tasks are not highly unbalanced in data size.

\myparagraph{Data Preprocessing}: SC and FSRL are in nature text classification and token classification tasks. Thus we use the raw texts from their datasets as inputs. NC, NAD, and TSA are text classification tasks, but they also require target companies or target numbers as inputs. Therefore, we use ``|COMPANY|'' to denote target companies and ``<NUMBER>'' to denote target numbers in input texts. CD is originally a span prediction task. For simplicity, we model it as a token classification task by converting the span labels to BIO tags (i.e., beginning and ending cause/effect spans to ``B-CAUSE I-CAUSE...'' and ``B-EFFECT I-EFFECT...'').

\myparagraph{Hyperparameters}: All models are fine-tuned with a initial learning rate of $0.00005$, warm up steps of $500$, and weight decay of $0.01$. Batches sizes we used for TSA, SC, NC, NAD, FSRL, and CD are $16$, $16$, $24$, $32$, $16$, and $16$ correspondingly. For the prediction header, we use a single feed-forward layer followed by Softmax.

\begin{table*}[t]
\centering
\small
\begin{tabular}{l|ccccc}
\hline
Metric            & BERT cased                       & BERT uncased                     & Y-FinBERT                        & FinancialBERT                    & P-FinBERT                        \\ \hline
Cosine Similarity$\uparrow$ & $0.8282{\scriptstyle \pm0.0095}$ & $0.8545{\scriptstyle \pm0.0039}$ & $0.8270{\scriptstyle \pm0.0067}$ & $0.7755{\scriptstyle \pm0.0166}$ & $0.8573{\scriptstyle \pm0.0057}$ \\
RMSE$\downarrow$            & $0.2320{\scriptstyle \pm0.0082}$ & $0.2069{\scriptstyle \pm0.0027}$ & $0.2275{\scriptstyle \pm0.0061}$ & $0.2500{\scriptstyle \pm0.0062}$ & $0.2054{\scriptstyle \pm0.0057}$ \\ \hline
\end{tabular}
\caption{\label{tab:metric_corr} TSA results obtained with different metrics for checkpoint selection and testing. The Pearson Correlation Coefficient between RMSE and Cosine Similarity is \textbf{$-0.9659$}, indicating that these metrics are approximately equivalent for TSA evaluation.}
%\vspace{-1.0em}
\end{table*}
\myparagraph{Evaluation Metrics Selection and Reporting}: The evaluation metrics are used not only for testing but also for best checkpoint selection during validation. Therefore, we report single metrics for all results to reflect MTL's effect on each task. We choose Accuracy for SC, NC, and NAD since these datasets have no severe label imbalance. For simplicity, we equivalently model CD, which is originally a span prediction task, as a token classification task, and use Accuracy as the metric. TSA is officially measured with cosine similarity \citep{cortis-etal-2017-semeval}. We find RMSE, as a regular metric for regression tasks, has a high correlation with cosine similarity (see \cref{tab:metric_corr}). Therefore, RMSE is suitable for TSA measurement. Besides, we avoid reporting average scores across tasks like related work because it makes no sense to average RMSE with Accuracy and F1 scores. 

\myparagraph{Evaluation Tools}: We use sklearn 1.0.2 for sequence classification evaluation, and seqeval 1.2.2 for token classification evaluation.

\myparagraph{GPU Usage}: Experiments are trained on NVIDIA RTX2080 GPUs. A single run of STL experiments takes 4 to 16 GPU hours (4 GPU hours for the small datasets; 16 GPU hours for the large ones). A single run of MTL experiments takes 16 to 96 GPU hours (16 GPU hours for the smallest subsets of FinDATA, e.g., SC and TSA; 96 GPU hours for full FinDATA).
 
\section{Financial NLP Datasets} \label{sec:all_datasets}
The detailed information of Financial NLP datasets discussed in \cref{sec:fin_nlp} is shown in \cref{tab:all_stat}. We only cover English datasets, and include the English subset for those multilingual datasets (e.g., FNS and FinTOC). Most of the datasets have less than $10$K data points in total, with fewer samples for training. Some data sizes are even fewer than $2$K.
\begin{table*}[t]
\centering
\resizebox{\textwidth}{!}{
\begin{tabular}{llll}
\hline
\textbf{Dataset}                 & \textbf{Task}                            & \multicolumn{1}{l}{\textbf{|Total|}} & \textbf{Text Source}       \\ \hline
Financial PhraseBank \citep{financialphrasebank2013malo}    & sentiment classification        & 4,837                       & Financial news    \\
StockSen \citep{xing-etal-2020-financial}                & sentiment classification        & 20,675                      & Financial tweets  \\
SemEval-2017 task-5-1 \citep{cortis-etal-2017-semeval}   & target-based sentiment analysis & 2,510                       & Financial tweets  \\
SemEval-2017 task-5-2 \citep{cortis-etal-2017-semeval}   & target-based sentiment analysis & 1,647                       & Financial news    \\
FNS \citep{el-haj-etal-2020-financial}               & Summarization                   & 12,796                      & UK annual report  \\
FinQA \citep{chen-etal-2021-finqa}                  & Numeracy question answering     & 8,281                       & Earning reports   \\
TAT-QA \citep{zhu-etal-2021-tat}                 & Tabular question answering      & 16,552                      & Financial reports \\
FinNum-1 \citep{chen2019overview}                & Number classification           & 8,868                       & Financial tweets  \\
FinNum-2 \citep{chen2020overview}               & Number attachement              & 10,340                      & Financial tweets  \\
FinNum-3 \citep{chen2022overview}               & Number classification           & 9,528                       & Analyst reports   \\
Numeracy-600K subtask-1 \citep{chen-etal-2019-numeracy} & Number magnitude prediction     & 600,000                     & Market comments   \\
Numeracy-600K subtask-2 \citep{chen-etal-2019-numeracy} & Number magnitude prediction     & 600,000                     & Financial news    \\
TAP \citep{lamm-etal-2018-textual}                    & Quantitative SRL                & 1,100                        & Financial news    \\
FinCausal \citep{mariko-etal-2020-financial}              & Causal effect detection         & 1,126                      & Financial news    \\
FinSim-2 \citep{maarouf-etal-2020-finsim}               & Financial concept understanding & 199 (concepts)                         & -                 \\
FinSim-3 \citep{kang-etal-2021-finsim}               & Financial concept understanding & 1,394 (concepts)                       & -                 \\ 
FinTOC \citep{maarouf-etal-2021-financial}               & TOC extraction & 72 (documents)                       & Financial prospectuses \\ \hline
\end{tabular}
}
\caption{\label{tab:all_stat} Detailed information of the existing Financial NLP datasets. We report the total number of samples in all dataset splits (train, test, and development). The text source of each dataset is also reported except for two concept-based datasets.}
\end{table*}

\section{Dataset Splits} \label{sec:data_split}
For FinDATA tasks, we use the official test set and development set if they exist and are publicly available: for TSA and FSRL, we use official test sets; for NC and NAD, we use both official test and development sets. If there is no available official test or development set, we split the datasets with a random seed of $42$: for SC, we split $10\%$ for test and $10\%$ for validation. For TSA, we split $20\%$ for validation. For FSRL, we split $10\%$ for validation. For CD, we split $20\%$ for test and $20\%$ for validation. All these data splits are available in our GitHub repository.

\section{Gradient Analysis} \label{sec:gradient_analysis}
We are curious whether gradient similarities reflect task-relatedness. Furthermore, do gradient conflicts/similarities interpret why some task aggregation works better than others? During MTL, we record each task's gradient (averaged over the whole training set) every 2000 training steps. Then we calculate the pairwise cosine similarity between the gradients of all task pairs. 

\begin{figure}[t]
    \centering
    \includegraphics[width=\columnwidth]{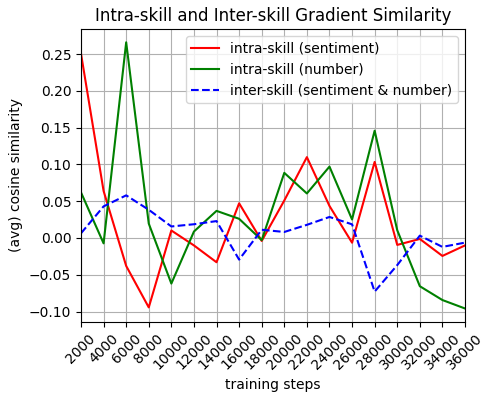}
    \caption{Intra-skill and inter-skill average gradient cosine similarity. All gradient similarities are measured on an MTL system including all FinDATA tasks (with a random seed of 1).}
    %\vspace{-0.7em}
    \label{fig:gc_skill}
\end{figure}

\myparagraph{Gradient similarity fails to reflect task-relatedness}: \cref{fig:gc_skill} shows the gradient similarity within sentiment and number tasks (intra-skill gradient similarity), and the average pairwise gradient similarity in-between the sentiment and number tasks (inter-skill gradient similarity). It can be observed that intra-skill gradients are not significantly more similar than inter-skill gradients, indicating that gradient similarity might not be a good measurement for task-relatedness. 

\begin{figure}[t]
    \centering
    \includegraphics[width=\columnwidth]{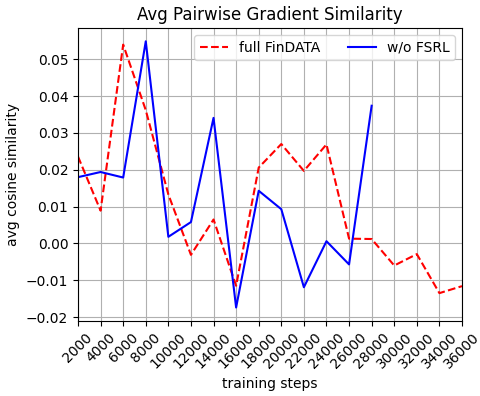}
    \caption{Average gradient similarities of two MTL systems: full FinDATA and ablating FSRL. Both are trained for 40 epochs, recording gradients every 2000 steps.}
    %\vspace{-0.7em}
    \label{fig:gc_all}
\end{figure}
\myparagraph{Gradient similarity does not indicate transferability within task aggregation}: \cref{fig:gc_all} shows the average pairwise gradient similarity of two MTL systems with different task aggregation: one is trained on full FinDATA, and the other ablates FSRL. Although ablating FSRL leads to worse scores on all tasks (see \cref{tab:mtl}), the gradient conflict of ``w/o FSRL'' is not significantly higher than full FinDATA. Therefore, gradient conflicts/similarities are not a good indicator of task aggregation quality.

\section{MTL Pre-training \& Unseen task Generalization} \label{sec:generalization}
\begin{table*}[t]
\centering
\small
\begin{tabular}{l|lclclc}
\hline
\multirow{2}{*}{Method} &  & \multicolumn{3}{c}{Numeracy-600K few-shot}      &  & StockSen few-shot        \\ \cline{3-3} \cline{5-5} \cline{7-7} 
                                 &  & Market Comment   &  & News Headline    &  & -                \\ \hline
FinBERT (Vanilla)                         &  & $18.34{\scriptstyle \pm1.1}$ &  & $14.25{\scriptstyle \pm2.5}$           &  & $58.33{\scriptstyle \pm4.0}$          \\
FinBERT (MTL)                      &  & $\mathbf{20.79}{\scriptstyle \pm0.6}$          &  & $\mathbf{20.56}{\scriptstyle \pm2.8}$          &  & $\mathbf{62.35}{\scriptstyle \pm5.2}$          \\ \hline
%SPAL-FinBERT (MTL)                 &  & $17.29{\scriptstyle \pm1.2}$          &  & $16.19{\scriptstyle \pm3.2}$ &  & $\mathbf{65.06}{\scriptstyle \pm3.6}$ \\ \hline
\end{tabular}
\caption{\label{tab:gene} Few-shot generalization results, where FinBERT (Vanilla) denotes FinBERT without MTL; and FinBERT (MTL) denotes systems with MTL pre-training before few-shot generalization. \textbf{Bold} number denotes the highest generalization score for each task. For Numeracy-600K we report Macro-F1 due to the imbalanced test set. For StockSen, we report accuracy. 
}
%\vspace{-1.5em}
\end{table*}
MTL pre-training may increase the model's generalizability to unseen tasks \cite{aghajanyan-etal-2021-muppet,karimi-mahabadi-etal-2021-parameter,ponti2022combining}, which might be extremely helpful when there is a shortage in target training data (a few-shot setting). Therefore, we test the few-shot generalizability of our MTL systems on two unseen tasks: StockSen and Numeracy-600K. StockSen is a binary (positive or negative) sentiment classification dataset on financial tweets. Numeracy-600K classifies numbers into one of seven magnitudes. It has two subtasks on different domains (financial news and market comment). We first train the models on FinDATA for $2000$ steps. Then we resume the shared encoder and fine-tune it on the target unseen task for $10$ epochs, reporting the best checkpoint's score. We use a few-shot setting (randomly sample $400$ training and $400$ validation data points) for unseen tasks to stimulate the lack of training data in the target task. For test sets, we split (with a random seed of $42$) $60$K samples ($10\%$ of data) for Numeracy-600K and $6.2$K samples (official development set) for StockSen. The results are shown in \cref{tab:gene}. In all tasks, the MTL-pre-trained system beats vanilla FinBERT when generalizing to unseen tasks. 
% It can also be seen that the absence of core skill during pre-training leads to worse generalizability to the target task. For Numeracy-600K, the best scores are obtained with the full FinDATA, indicating that skill diversity is essential when generalizing to a new task (e.g. number magnitude detection). For StockSen, the best score is obtained by concentrating on sentiment tasks. A possible explanation is that sentiment analysis tasks are very similar to each other thus benefits from concentration. SPAL-FinBERT pre-trained on full FinDATA also generalizes well to StockSen ($65.06$ which is slightly lower than the best score). Therefore, generally speaking, MTL pre-training with larger scale benefits the model's generalizability more.

\section{Importance of Freezing Pretrained Model} \label{appendix:freeze}
To illustrate the importance of freezing the pre-trained model, we first compare SPAL-FinBERT (SPAL hidden size = 204) with an ablation setting where FinBERT is not frozen. The comparison is shown in \cref{tab:ablation}, where unfreezing FinBERT compromises the MTL performance drastically on most tasks (CD prefers larger shared capacity and thus benefits from unfreezing). 

\begin{figure}[t]
    \centering
    \includegraphics[width=\columnwidth]{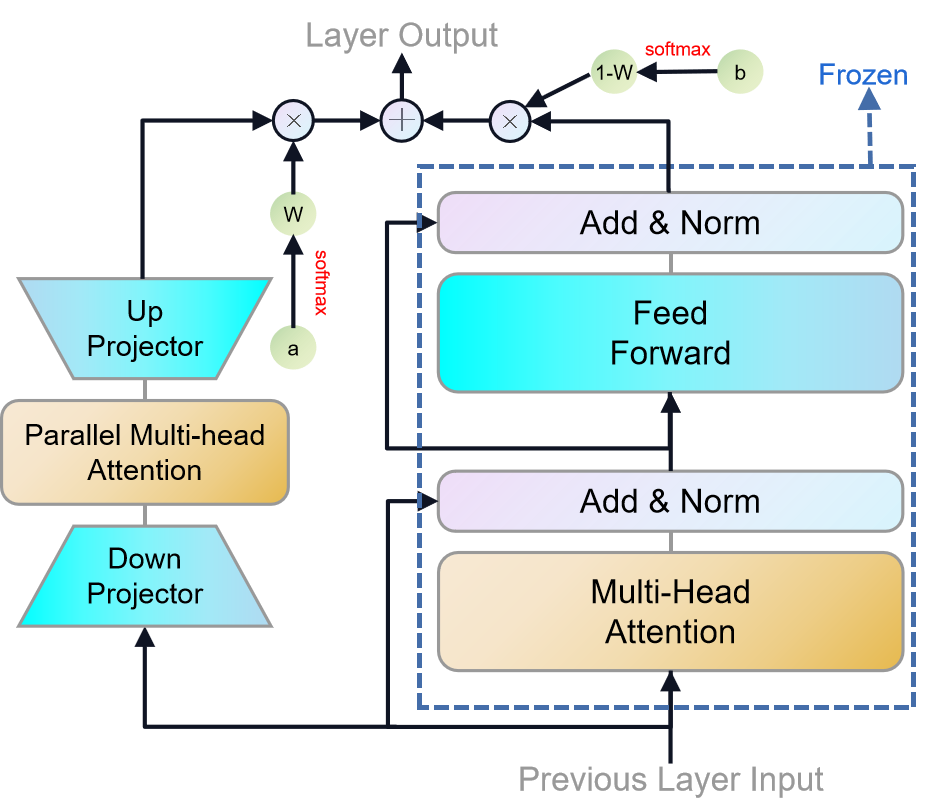}
    \caption{The contribution probing architecture where $a$ and $b$ denote the attention parameters; $w$ and $1-w$ denote the weights after softmax.}
    \vspace{-0.7em}
    \label{fig:prob}
\end{figure}

\begin{figure}[t]
    \centering
    \includegraphics[width=\columnwidth]{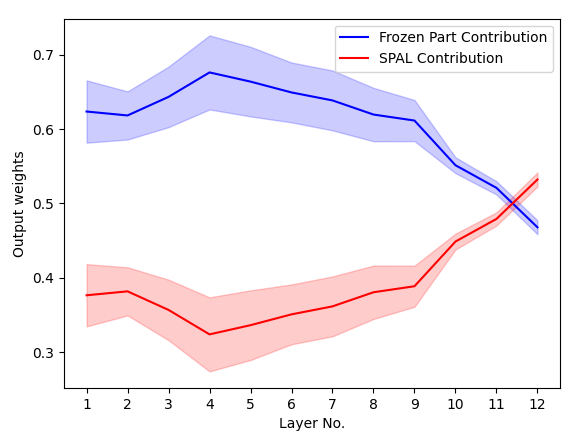}
    \caption{Contributions of the frozen part and the PAL to the layer output in each layer.}
    \vspace{-0.7em}
    \label{fig:attention_weights}
\end{figure}

Then we add weighting parameters to probe the frozen FinBERT's contribution to the layer outputs. \cref{fig:prob} shows the probing architecture. We add probing parameters $a$ and $b$, which weigh the frozen FinBERT output and the SPAL output. After MTL, the contribution of each structure can be measured by the final (softmaxed) weights. The results are shown in \cref{fig:attention_weights}. In all layers except the last layer, the frozen FinBERT layers contribute more to the output than PALs, illustrating the importance of the frozen part. \\
\begin{table*}[t]
\centering
\small
\begin{tabular}{l|llllll}
\hline
Task                                                                          & \multicolumn{1}{c}{TSA$\downarrow$}                              & \multicolumn{1}{c}{SC}                           & \multicolumn{1}{c}{NC}                           & \multicolumn{1}{c}{NAD}                          & \multicolumn{1}{c}{FSRL}                         & \multicolumn{1}{c}{CD}                           \\ \hline
\begin{tabular}[c]{@{}l@{}}SPAL-FinBERT\\ Multi-task w/o Freeze\end{tabular} & $0.2231{\scriptstyle \pm0.0075}$                     & $86.15{\scriptstyle \pm1.4}$                     & $87.10{\scriptstyle \pm0.5}$                     & $85.95{\scriptstyle \pm0.5}$                     & $70.50{\scriptstyle \pm1.2}$                     & $\mathbf{78.12}{\scriptstyle \pm1.0}$                     \\ \hline
\begin{tabular}[c]{@{}l@{}}SPAL-FinBERT\\ Multi-task\end{tabular}              & $\mathbf{0.2093}{\scriptstyle \pm0.0032}$                     & $\mathbf{87.10}{\scriptstyle \pm0.5}$                     & $\mathbf{87.82}{\scriptstyle \pm0.7}$                     & $\mathbf{86.35}{\scriptstyle \pm0.3}$                     & $\mathbf{71.92}{\scriptstyle \pm1.4}$                     & $74.88{\scriptstyle \pm0.7}$                     \\ \hline
\end{tabular}
\caption{\label{tab:ablation}
comparison between SPAL-FinBERT (SPAL hidden size = 204) with frozen and unfrozen FinBERT. Metrics reported for FinDATA tasks are the same as \cref{tab:mtl}.
}
%\vspace{-0.7em}
\end{table*}

\section{Task-relatedness Examples} \label{sec:transfer_example}

\begin{figure*}[t]
     \centering
     \begin{subfigure}[t]{0.47\textwidth}
         \centering
         \includegraphics[width=\textwidth]{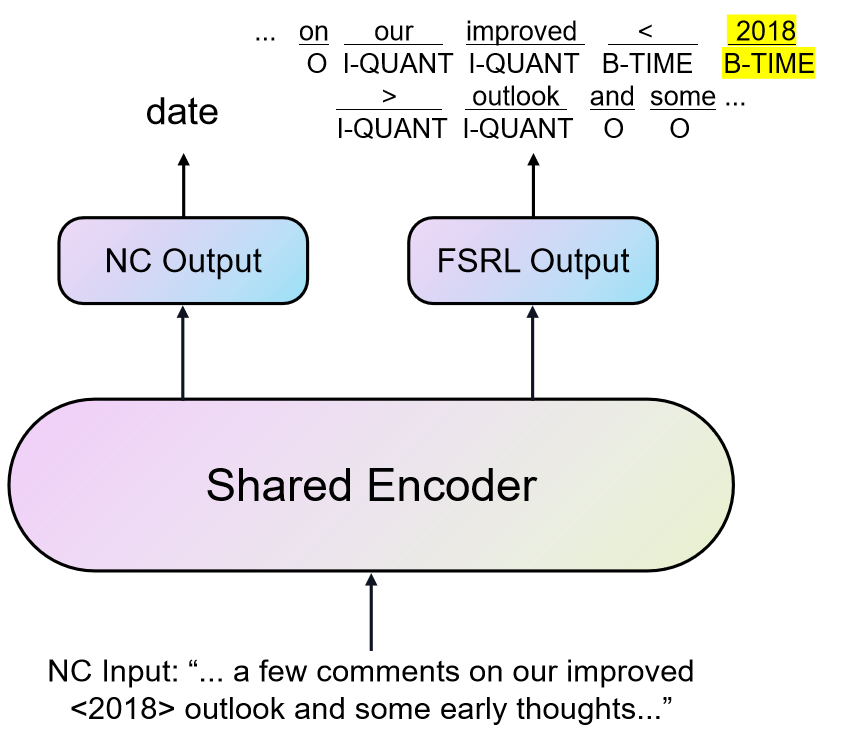}
         \caption{Input a sentence from NC test set, where the target number is 2018. FSRL (non-target) header's output shows that time awareness is injected.}
         \label{fig:skill_a}
     \end{subfigure}
     \hfill
     \begin{subfigure}[t]{0.47\textwidth}
         \centering
         \includegraphics[width=0.86\textwidth]{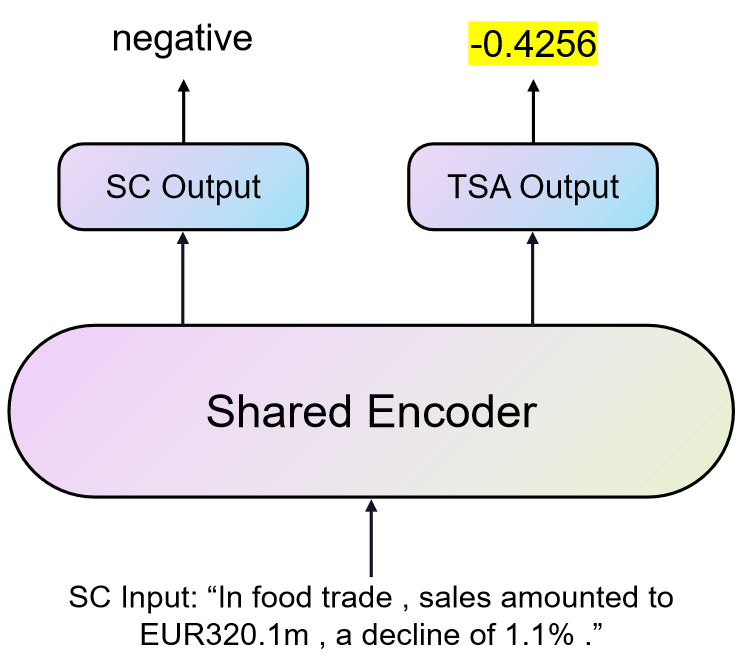}
         \caption{Input a sentence from SC test set, TSA (non-target) header's output shows that sentiment analysis skill is enhanced by TSA.}
         \label{fig:skill_b}
     \end{subfigure}
        \caption{Some explicit positive transfer examples.}
        \label{fig:three graphs}
\end{figure*}

Through MT fine-tuning, the shared encoder understands an input sentence from comprehensive aspects that positively transfer to each other. To probe the explicit transfer, we analyze the non-target output headers' outputs to illustrate that the inputs are understood comprehensively. For example, \cref{fig:skill_a} shows that the FSRL header correctly identifies the semantic role of ``2018'' in an NC input. Such time awareness may benefit NC when classifying the date numbers. Similarly, \cref{fig:skill_b} shows that the TSA header assigns proper sentiment score to an SC input. More examples of explicit transfer are shown below:

\noindent Positive transfer from TSA to SC (\textcolor{blue}{target header: SC}, \textcolor{red}{non-target header: TSA}). All examples are from SC test set:
\begin{itemize}
    \item Finnish Aldata Solution has signed a contract of supply its G.O.L.D. system to two French retail chains. \textcolor{blue}{Golden label: positive}; \textcolor{red}{TSA output: 0.40566313}
    \item Kaupthing Bank will publish its annual results for 2007 before markets open on Thursday 31 January. \textcolor{blue}{Golden label: neutral}; \textcolor{red}{TSA output: -0.00095879}
    \item In food trade , sales amounted to EUR320.1 m , a decline of 1.1\% . \textcolor{blue}{Golden label: negative}; \textcolor{red}{TSA output: -0.42561457}
    \item The company did not distribute a dividend in 2005. \textcolor{blue}{Golden label: neutral}; \textcolor{red}{TSA output: -0.35266992}
    \item Panostaja did not disclose the purchase price. \textcolor{blue}{Golden label: neutral}; \textcolor{red}{TSA output: 0.02710678}
    \item Operating profit rose to EUR 13.5 mn from EUR 9.7mn in the corresponding period in 2006. \textcolor{blue}{Golden label: positive}; \textcolor{red}{TSA output: 0.42477337}
    \item As production of other products will continue normally, temporary lay-offs concern simultaneously at most 80 employees. \textcolor{blue}{Golden label: negative}; \textcolor{red}{TSA output: -0.49556375}
    \item According to Viking Line's Managing Director, Nils-Erik Eklund, the company's Board of Directors is very satisfied with Viking Line's performance. \textcolor{blue}{Golden label: positive}; \textcolor{red}{TSA output: 0.28375068}
    \item The port operator, however, favors retaining the port fees in 2010, citing the owner, the government of Estonia, committing the port to pay EEK 400mn (EUR 25.56 mn USD 36.44 mn) in dividends to the state in 2009 and another EEK 300mn in 2010. \textcolor{blue}{Golden label: neutral}; \textcolor{red}{TSA output: 0.05767085}
    \item Uponor maintains its full-year guidance for 2010. \textcolor{blue}{Golden label: neutral}; \textcolor{red}{TSA output: 0.08272883}
\end{itemize}

\noindent Positive transfer from FSRL to NC (date classification). Semantic role labeling outputs are in a form of [(token, label)] entries. Time semantic roles are \textbf{bold}. All examples are from NC test set:
\begin{itemize}
    \item Looking ahead the 150000 people at Optum are incredibly enthusiastic about <2019> and our opportunities for longer-term growth and... \textcolor{red}{FSRL output}: [('looking', 'O'), ('ahead', 'O'), ('the', 'O'), ('150000', 'I-WHOLE'), ('people', 'I-WHOLE'), ('at', 'I-WHOLE'), ('optum', 'B-SOURCE'), ('are', 'O'), ('incredibly', 'O'), ('enthusiastic', 'O'), ('about', 'O'), \textbf{('<', 'B-TIME'), ('2019', 'B-TIME')}, ('>', 'I-QUANT'), ('and', 'O'), ('our', 'O'), ('opportunities', 'O'), ('for', 'O'), ('longer', 'O'), ('-', 'O'), ('term', 'O'), ('growth', 'O'), ('and', 'O') ...
    \item Now before I turn it over to Carroll just a few comments on our improved <2018> outlook and some early thoughts on 2019 ... \textcolor{red}{FSRL output}: [('now', 'B-TIME'), ('before', 'O'), ('i', 'O'), ('turn', 'O'), ('it', 'O'), ('over', 'O'), ('to', 'O'), ('carroll', 'O'), ('just', 'O'), ('a', 'O'), ('few', 'O'), ('comments', 'O'), ('on', 'O'), ('our', 'I-QUANT'), ('improved', 'I-QUANT'), \textbf{('<', 'B-TIME'), ('2018', 'B-TIME')}, ('>', 'I-QUANT'), ('outlook', 'I-QUANT'), ('and', 'O'), ('some', 'O'), ('early', 'O'), ('thoughts', 'I-QUANT'), ('on', 'O'), ('2019', 'I-QUANT') ...
    \item Now before I turn it over to Carroll just a few comments on our improved 2018 outlook and some early thoughts on <2019.> ... \textcolor{red}{FSRL output}: [('now', 'O'), ('before', 'O'), ('i', 'O'), ('turn', 'O'), ('it', 'O'), ('over', 'O'), ('to', 'O'), ('carroll', 'O'), ('just', 'O'), ('a', 'O'), ('few', 'O'), ('comments', 'O'), ('on', 'O'), ('our', 'I-QUANT'), ('improved', 'I-QUANT'), ('2018', 'I-QUANT'), ('outlook', 'I-QUANT'), ('and', 'O'), ('some', 'O'), ('early', 'O'), ('thoughts', 'I-QUANT'), ('on', 'I-QUANT'), \textbf{('<', 'I-TIME'), ('2019', 'I-TIME'), ('.', 'I-TIME'), ('>', 'I-TIME')}...
\end{itemize}

\section{FinDATA Examples} \label{sec:findata_examples}
In this section we provide $10$ examples for each FinDATA task:

\noindent \textbf{TSA}: \citep{cortis-etal-2017-semeval} the target companies are enclosed by ``| |'':
\begin{itemize}
    \item NYSE owner |ICE| considers offer for LSE. \textcolor{red}{Label: 0.096}
    \item NYSE owner ICE considers offer for |LSE|. \textcolor{red}{Label: 0.396}
    \item |Diageo| sales disappoint as currency and comparatives leave bitter taste. \textcolor{red}{Label: -0.545}
    \item AB InBev attacks |SABMiller| bid rebuffal. \textcolor{red}{Label: -0.158}
    \item Are ARM Holdings plc, |Domino's Pizza Group plc| and ASOS plc 3 must-have growth stocks?. \textcolor{red}{Label: 0.063}
    \item Drugmaker |Shire| to buy Baxalta for \$32 billion after 6-month pursuit. \textcolor{red}{Label: 0.437}
    \item Drugmaker Shire to buy |Baxalta| for \$32 billion after 6-month pursuit. \textcolor{red}{Label: 0.75}
    \item Centrica extends gas deals with Gazprom, |Statoil|. \textcolor{red}{Label: 0.239}
    \item |Aggreko| 2015 Profit Declines - Quick Facts. \textcolor{red}{Label: -0.441}
    \item |HSBC| shakes up board with two new business chiefs, three departures. \textcolor{red}{Label: -0.074}
\end{itemize}
\noindent \textbf{SC} \citep{financialphrasebank2013malo}: 
\begin{itemize}
    \item The business to be divested generates consolidated net sales of EUR 60 million annually and currently has some 640 employees. \textcolor{red}{Label: neutral}
    \item Svyturys-Utenos Alus, which is controlled by the Nordic group Baltic Beverages Holding (BBH), posted a 6.1 percent growth in beer sales for January-September to 101.99 million liters. \textcolor{red}{Label: positive}
    \item The Department Store Division's sales fell by 8.6\% to EUR 140.2 mn. \textcolor{red}{Label: negative}
    \item Production capacity will rise gradually from 170,000 tonnes to 215,000 tonnes. \textcolor{red}{Label: positive}
    \item Rautalinko was resposnible also for Mobility Services, and his job in this division will be continued by Marek Hintze. \textcolor{red}{Label: neutral}
    \item Circulation revenue has increased by 5\% in Finland and 4\% in Sweden in 2008. \textcolor{red}{Label: positive}
    \item The changes will take effect on 1 January 2010, and they are not estimated to have an impact on the number of employees. \textcolor{red}{Label: neutral}
    \item F-Secure Internet Security 2010 is a security service for surfing the web, online banking and shopping, e-mail, and other online activities. \textcolor{red}{Label: neutral}
    \item Earnings per share (EPS) were EUR0.03, up from the loss of EUR0.083. \textcolor{red}{Label: positive}
    \item Production capacity will increase from 36000 to 85000 tonnes per year and the raw material will continue to be recycled paper and board. \textcolor{red}{Label: positive}
\end{itemize}
\noindent \textbf{NC}: \citep{chen2022overview} the targeted numbers are enclosed by ``\textless{} \textgreater{}'':
\begin{itemize}
    \item Finally we experienced roughly \$<104> million of hurricane-related expenses in the quarter for items like people-cost increased security in our affected stores and storm damage. So while our year-over-year sales growth was positively impacted by the hurricanes our operating profit was negatively impacted by \$51 million. \textcolor{red}{Label: money}
    \item In Asia we expect to acquire 51\% of our Philippines bottler from Coca-Cola FEMSA during the fourth quarter. This will become a part of our Bottling Investments Group which is now comprised primarily of Southwest and Southeast Asian bottlers. These \textless{}2\textgreater transactions should roughly offset each other resulting in a minimal structural impact in our P\&L in 2019. \textcolor{red}{Label: money} \textcolor{red}{Label: quantity\_absolute}
    \item From a capital allocation perspective year-to-date we have generated \$6.3 billion of free cash flow returned \$\textless{}8.6\textgreater billion to shareholders including \$2.8 billion in dividends and \$5.8 billion in buybacks repurchasing 117 million shares. \textcolor{red}{Label: money}
    \item Next on Aviation which had another great quarter. Orders of \$8.8 billion were up 12\%. Equipment orders grew 20\% driven by the continued strong momentum of the LEAP engine program up 56\% versus the prior year. Military engine orders were up 69\% driven by the F414 and service orders grew 7\%. Revenues of \$8.5 billion grew \textless{}21\textgreater{}\%. Equipment revenues were up 13\% on higher commercial engines partially offset by lower military volume. \textcolor{red}{Label: relative}
    \item We'll release 2 new movies from Pixar in fiscal \textless{}2018\textgreater{}. We're thrilled with the early reaction to Coco which opens at Thanksgiving and we're also looking forward to the summer release of The Incredibles 2. \textcolor{red}{Label: date}
    \item I would like to remind you that some of the statements that we make during today's call may be considered forward-looking statements within the meaning of the safe harbor provision of the U.S. Private Securities Litigation Reform Act of \textless{}1995\textgreater{}. \textcolor{red}{Label: other}
    \item We ended 2017 with franchised restaurants representing \textless{}92\textgreater{}\% of our total restaurant base up from 81\% 3 years ago. As a result franchise margins now comprise more than 80\% of our total restaurant margin dollars. For the fourth quarter franchise margin dollars increased across all segments reflecting sales-driven performance and the shift to a more heavily franchised system. \textcolor{red}{Label: absolute}
    \item Today we announced that we will increase our quarterly dividend by 15\% or by \$0.07 to \$0.55 per share beginning in the first quarter of 2019. In addition, the board has approved an additional \$<10> billion share repurchase authorization giving us approximately \$18 billion in share repurchase capacity. \textcolor{red}{Label: money}
    \item Looking back on 2017 I could not be more proud of our team and all they have accomplished. As I look to our \textless{}50\textgreater{}th year I'm more optimistic and confident than I've ever been about Intel's future. \textcolor{red}{Label: quantity\_absolute}
    \item Non-GAAP gross margin was \textless{}76\textgreater{}\% in the quarter an increase of roughly 70 basis points versus the third quarter of 2016. Favorable product mix driven by KEYTRUDA and ZEPATIER was the largest contributor to the year-over-year improvement. Non-GAAP operating expenses of \$4.2 billion increased 4\% year-over-year primarily driven by higher R\&D expense reflecting increased investments in early drug development. Taken together we earned \$1.11 per share on a non-GAAP basis up 4\% excluding exchange. Note that our GAAP EPS loss of \$0.02 reflects the charge of \$2.35 billion related to the formation of the strategic oncology collaboration with AstraZeneca announced earlier in the quarter. \textcolor{red}{Label: absolute}
\end{itemize}
\noindent \textbf{NAD}: \citep{chen2020overview} target numbers and cash tags are indicated by ``\textless{} \textgreater{}'' and ``| |'' correspondingly:
\begin{itemize}
    \item \$|XXII| Scott Gottlieb, Commissioner of FDA speech transcript from November \textless{}3\textgreater{}rd, less than 2 months left in year then. \textcolor{red}{Label: attached}
    \item \$|DPW| that was quite a roller coaster. Glad it ended well. Should see \textless{}5\textgreater in 7 days \textcolor{red}{Label: attached}
    \item Took me \textless{}5\textgreater minutes to conclude: \#Snoozefest \textbackslash{}ud83d \textbackslash{}ude34\textbackslash{}ud83d \textbackslash{}ude34 \textbackslash{}ud83d \textbackslash{}ude34 \textbackslash{}ud83d \textbackslash{}ude34 Advancers 6 to Decliners 5 NYSE + NASDAQ \$|SPY| \$QQQ \$DIA \$IWM \textcolor{red}{Label: unattached}
    \item Take moment \textless{}2\textgreater note \$Crypto Superiority trades 24/7 365 No dead time 4 Thanksgiving \$BTC.X \$LTC.X \$|ETH.X| \$DASH.X \$XRP.X \$BCH.X \$TSLA \$MNKD \textcolor{red}{Label: unattached}
    \item 2nd TP for \$|JDST| is 94.05 according to my algo. Take it to the bank. Gold headed <4> major intermediate bottom b4 spike in Jan 2018. \$JNUG \textcolor{red}{Label: unattached}
    \item 2nd TP for \$|JDST| is 94.05 according to my algo. Take it to the bank. Gold headed 4 major intermediate bottom b4 spike in Jan |2018|. \$JNUG \textcolor{red}{Label: unattached}
    \item \$|BABA| hit +\$\textless{}3\textgreater pre-market - Futures up 100 - \$BSTI Big Buying @ close after Rebound Holds 2nd day Heading Back to \$13 \textcolor{red}{Label: attached}
    \item \$|BABA| hit +\$3 pre-market - Futures up \textless{}100\textgreater - \$BSTI Big Buying @ close after Rebound Holds 2nd day Heading Back to \$13 \textcolor{red}{Label: attached}
    \item \$|BABA| hit +\$3 pre-market - Futures up 100 - \$BSTI Big Buying @ close after Rebound Holds <2>nd day Heading Back to \$13 \textcolor{red}{Label: attached}
    \item \$|BABA| hit +\$3 pre-market - Futures up 100 - \$BSTI Big Buying @ close after Rebound Holds 2nd day Heading Back to \$\textless{}13\textgreater{}  \textcolor{red}{Label: attached}
\end{itemize}
\noindent \textbf{FSRL}: \citep{lamm-etal-2018-textual} we use different colors to denote different semantic roles: \textcolor{purple}{purple} for WHOLE, \textcolor{red}{red} for THEME, \textcolor{blue}{blue} for MANNER, \textcolor{forestgreen}{forestgreen} for VALUE, \textcolor{orange}{orange} for TIME, \textcolor{goldenrod}{goldenrod} for QUANT, \textcolor{pink}{pink} for AGENT, \textcolor{cyan}{cyan} for SOURCE, and \textcolor{sepia}{sepia} for CAUSE. For a detailed definition of each semantic role, please refer to \citet{Lamm2018QSRLA}.
\begin{itemize}
    \item \textcolor{purple}{Commodities}: \textcolor{red}{Dow Jones futures index} \textcolor{forestgreen}{129.72}, \textcolor{blue}{off} \textcolor{forestgreen}{0.15}; \textcolor{red}{spot index} \textcolor{forestgreen}{130.16}, \textcolor{forestgreen}{up} \textcolor{forestgreen}{0.91}.
    \item Between \textcolor{forestgreen}{50\% and 75\%} of \textcolor{orange}{today}'s \textcolor{purple}{workers} \textcolor{goldenrod}{are covered by such plans}, up from \textcolor{forestgreen}{5\%} \textcolor{orange}{five years ago}.
    \item \textcolor{pink}{Cary Computer}, which \textcolor{orange}{currently} \textcolor{goldenrod}{employs} \textcolor{forestgreen}{241 people}, said \textcolor{pink}{it} expexts a \textcolor{goldenrod}{work force} of \textcolor{forestgreen}{450} by \textcolor{orange}{the end of 1990}.
    \item \textcolor{red}{Colgate-Palmolive} advanced 1 5/8 to 63 after saying it was comfortable with \textcolor{cyan}{analysts}' projections that \textcolor{orange}{third-quarter} \textcolor{goldenrod}{net income from continuing operations} would be between \textcolor{forestgreen}{95 cents and \$1.05} a share, up from \textcolor{forestgreen}{69 cents} \textcolor{orange}{a year ago}.
    \item In addition, \textcolor{red}{CMS} reported \textcolor{orange}{third-quarter} \textcolor{goldenrod}{net} of \textcolor{forestgreen}{\$68.2 million}, or \textcolor{forestgreen}{83 cents a share}, up from \textcolor{forestgreen}{\$66.8 million}, or \textcolor{forestgreen}{81 cents a share}, \textcolor{orange}{a year ago}.
    \item \textcolor{purple}{Chateau Yquem}, the leading Sauternes, now \textcolor{goldenrod}{goes for well} over \textcolor{forestgreen}{\$100 a bottle} for \textcolor{red}{a lighter vintage} like 1984; the spectacularly rich \textcolor{red}{1983} \textcolor{goldenrod}{runs} \textcolor{forestgreen}{\$179}.
    \item For \textcolor{orange}{the nine months}, \textcolor{red}{Arco} reported \textcolor{goldenrod}{net income} of \textcolor{forestgreen}{\$1.6 billion}, or \textcolor{forestgreen}{\$8.87 a share}, up 33\% from \textcolor{forestgreen}{\$1.2 billion}, or \textcolor{forestgreen}{\$6.56 a share} \textcolor{orange}{a year earlier}.
    \item Citing \textcolor{sepia}{its reduced ownership in the Lyondell Petrochemical Co.}, \textcolor{red}{Atlantic Richfield} reported that \textcolor{goldenrod}{net income} slid 3.1\% in \textcolor{orange}{the third quarter} to \textcolor{forestgreen}{\$379 million}, or \textcolor{forestgreen}{\$2.19 a share}, from \textcolor{forestgreen}{\$391 million}, or \textcolor{forestgreen}{\$2.17 a share}, for \textcolor{orange}{the comparable period last year}.
    \item \textcolor{orange}{Quarter} \textcolor{goldenrod}{revenue} was \textcolor{forestgreen}{\$232.6 million}, up 12\% from \textcolor{forestgreen}{\$206 million} \textcolor{orange}{last year}.
    \item \textcolor{purple}{Life insurers} fared similarly, with \textcolor{red}{Legal \& General} \textcolor{blue}{advancing} \textcolor{forestgreen}{3} to \textcolor{forestgreen}{344}, although \textcolor{red}{Prudential} \textcolor{blue}{fell} \textcolor{forestgreen}{2} to \textcolor{forestgreen}{184 1/2}.
\end{itemize}
\noindent \textbf{CD}: \citep{mariko-etal-2020-financial} we use \textcolor{blue}{blue} to denote causes, and \textcolor{red}{red} to denote effects:
\begin{itemize}
    \item \textcolor{blue}{Florida is unique in that it also draws a large proportion of higher net-worth individuals}  -  \textcolor{red}{more than 85 percent of its net inflow of income came from people earning at least six-figures}.
    \item \textcolor{blue}{CLICK HERE TO GET THE FOX BUSINESS APP  Data from the U.S. Census Bureau showed that while Florida received more movers than any other state last year}, \textcolor{red}{New York's outflows to the Sunshine State were the highest  -  63,772 people.}
    \item \textcolor{red}{New York had the third-largest outflows of any state, with 452,580 people moving out within the past year.} \textcolor{blue}{Individuals earning \$650,000 can save more than \$69,700 in taxes per year by moving from New York to Florida.}
    \item \textcolor{blue}{The stock increased 1.02\% or \$0.23 during the last trading session}, \textcolor{red}{reaching \$22.69.}
    \item \textcolor{red}{(NASDAQ:SBRA) has declined 1.62\% since September 21, 2018 and is downtrending.} \textcolor{blue}{It has underperformed by 1.62\% the S\&P500.}
    \item \textcolor{red}{Weyerhaeuser Company (NYSE:WY) has declined 25.53\% since September 21, 2018 and is downtrending.} \textcolor{blue}{It has underperformed by 25.53\% the S\&P500.}
    \item \textcolor{blue}{After \$0.46 actual EPS reported by Sabra Health Care REIT, Inc. for the previous quarter}, \textcolor{red}{Wall Street now forecasts 2.17\% EPS growth.}
    \item \textcolor{blue}{Investors sentiment increased to 1.25 in Q2 2019.} Its up 0.38, from 0.87 in 2019Q1. \textcolor{red}{It increased, as 23 investors sold SBRA shares while 68 reduced holdings.}
    \item \textcolor{blue}{It also reduced its holding in Qualcomm Inc.} \textcolor{red}{(NASDAQ:QCOM) by 24,294 shares in the quarter, leaving it with 158,167 shares, and cut its stake in Wells Fargo\& Co (New) (NYSE:WFC).}
    \item \textcolor{blue}{Investors sentiment decreased to 1.02 in 2019 Q2.} Its down 0.11, from 1.13 in 2019Q1. \textcolor{red}{It worsened, as 43 investors sold WY shares while 242 reduced holdings.}
\end{itemize}

\end{document}